%% file: main.tex
\documentclass{article}
\PassOptionsToPackage{numbers,sort&compress}{natbib}
\usepackage[main,final]{neurips_2026}
\makeatletter
\renewcommand{\@notice}{}
\makeatother
\usepackage[T1]{fontenc}
\usepackage[utf8]{inputenc}
\usepackage{microtype}
\usepackage{amsmath,amssymb,mathtools}
\usepackage{empheq}
\usepackage{bm}
\usepackage{graphicx}
\usepackage{booktabs}
\usepackage[table]{xcolor}
\usepackage{tcolorbox}
\usepackage{siunitx}
\tcbuselibrary{theorems}
\definecolor{CasalBrand}{HTML}{35636E}
\definecolor{SiennaBrand}{HTML}{EB684C}
\definecolor{DeployedAccent}{HTML}{0072BB}
\definecolor{ProjectorAccent}{HTML}{6667AB}

\usepackage{hyperref}
\usepackage{url}
\usepackage[nameinlink,capitalise]{cleveref}
\crefname{equation}{Equation}{Equations}
\Crefname{equation}{Equation}{Equations}
\crefname{figure}{Figure}{Figures}
\Crefname{figure}{Figure}{Figures}
\crefname{table}{Table}{Tables}
\Crefname{table}{Table}{Tables}
\crefname{section}{Section}{Sections}
\Crefname{section}{Section}{Sections}
\crefname{subsection}{Section}{Sections}
\Crefname{subsection}{Section}{Sections}
\crefname{appendix}{Appendix}{Appendices}
\Crefname{appendix}{Appendix}{Appendices}

\usepackage{tikz}
\usetikzlibrary{arrows.meta,positioning}
\usepackage{tikz-cd}

\usepackage{pgfplots}
\usepackage{pgfplotstable}
\usepgfplotslibrary{fillbetween}
\pgfplotsset{compat=1.18}

\newcommand{\Winder}{\textsc{Winder}}

\newcommand{\code}[1]{\texttt{#1}}
\makeatletter
\newcommand{\nonfloatcaption}[1]{%
  \refstepcounter{figure}%
  \@makecaption{\figurename~\thefigure}{#1}%
}
\let\WinderOriginalMakeCaption\@makecaption
\long\def\@makecaption#1#2{%
  \begingroup
  \footnotesize
  \WinderOriginalMakeCaption{#1}{#2}%
  \endgroup
}
\def\ps@winderfirstpage{%
  \let\@oddhead\@empty
  \let\@evenhead\@empty
  \def\@oddfoot{%
    \vbox{%
      \hrule width 12pc height 0.4pt
      \vskip 3pt
      \hbox to \textwidth{%
        \footnotesize $^{*}$ Equal contribution; joint first authors.\hfil
      }%
    }%
  }%
  \let\@evenfoot\@oddfoot
}
\makeatother

\title{Capturing Cardiac Cyclicity through Phase-Equivariant Self-Supervised Learning}

\author{%
  Blaise Delaney$^{*}$ \\
  TimeTrace Labs
  \And
  Dominic Dootson$^{*}$ \\
  TimeTrace Labs
  \And
  Juan Jose Juan Castella \\
  TimeTrace Labs \\
  University of Cambridge
  \And
  Salil Patel \\
  TimeTrace Labs \\
  University of Oxford
  \And
  Andrew Pfaff \\
  TimeTrace Labs
  \And
  Yuji Xing \\
  TimeTrace Labs
  \And
  Jonny Hancox \\
  NVIDIA
  \And
  Karin Sevegnani \\
  NVIDIA
}

\begin{document}
\twocolumn[{%
  \maketitle
  \begin{abstract}
The cyclic structure of physiological processes offers a natural prior for self-supervised
representation learning, and the cardiac cycle provides a particularly well-defined setting in
which to exploit it. We derive a phase-equivariant self-supervised objective and introduce
\Winder{}, a joint-embedding architecture that organises representations into phase-invariant
coordinates and phase-rotating harmonic subspaces. Its transport operator is fixed and
closed-form, derived from the cycle's geometry rather than learned, and adds no parameters. Evaluated on PTB-XL under a frozen linear-probe protocol, \Winder{} attains diagnostic accuracy within the range reported by state-of-the-art self-supervised methods at a $\mathord{\sim}1\,\mathrm{M}$ parameter footprint, while exhibiting phase-equivariant latent geometry. These findings demonstrate that explicitly encoding cardiac-phase symmetry can preserve diagnostically useful information while yielding a latent geometry that is legible, parameter-efficient, and directly tied to a measurable physiological quantity.
  \end{abstract}
  \input{phase-wrap-figure}
}]
\thispagestyle{winderfirstpage}

\input{01_introduction}
\input{02_related_work}
\input{03_theory}
\input{method}
\input{04_results}

\input{05_discussion}
\input{06_conclusions}

\begin{ack}
We thank Thorir Mar Ingolfsson for helpful discussions in the preparation of this work.
The authors acknowledge the use of resources provided by the Isambard 3 Tier-2 HPC Facility. Isambard 3 is hosted by the University of Bristol and operated by the \href{https://gw4.ac.uk}{GW4 Alliance} and is funded by UK Research and Innovation; and the Engineering and Physical Sciences Research Council [EP/X039137/1]. This work was supported by Innovate UK, part of UK Research and Innovation (UKRI), under grant number 10201397.
\end{ack}

\bibliographystyle{unsrtnat}
\bibliography{winder,ptbxl}

\appendix
\input{08_appendix_data}
\input{09_appendix_phase_assignment}
\input{training_appendix}
\input{07_appendix_ptbxl}
\end{document}

%% file: phase-wrap-figure.tex
\begin{center}
    \begin{minipage}[c]{0.53\textwidth}
        \begin{tikzpicture}[remember picture]
            \begin{axis}[
                width=\linewidth,
                height=9.5cm,
                axis lines=left,
                axis y line=none,
                axis x line=bottom,
                axis line style={
                    -{Triangle[length=3mm,width=2.2mm]},
                    line width=1.0pt
                },
                xmin=-0.17,
                xmax=1.01,
                ymin=-0.23,
                ymax=1.15,
                xtick={0,0.84},
                xticklabels={$R_i$,$R_{i+1}$},
                xlabel={Time, $t\in\mathbb{R}$},
                xlabel style={
                    name=time-label,
                    font=\large,
                    yshift=6mm
                },
                clip=false,
            ]
                \draw[
                    gray!70,
                    densely dashed,
                    line width=0.8pt
                ]
                    (axis cs:0,-0.23)
                    --
                    (axis cs:0,0.4491174561);
                \draw[
                    gray!70,
                    densely dashed,
                    line width=0.8pt
                ]
                    (axis cs:0.84,-0.225)
                    --
                    (axis cs:0.84,0.4491174561);
                \begin{scope}[overlay]
                \addplot[
                    name path=linear-curtain-top,
                    draw=none,
                    restrict x to domain=0:0.84,
                ]
                table[
                    col sep=comma,
                    x=time_s,
                    y=amplitude,
                ]{data/phase_wrap_time_compressed.csv};
                \path[name path=linear-curtain-bottom]
                    (axis cs:0,-0.23)
                    --
                    (axis cs:0.84,-0.23);
                \addplot[
                    fill=CasalBrand,
                    fill opacity=0.12,
                    draw=none,
                ]
                fill between[
                    of=linear-curtain-top and linear-curtain-bottom
                ];
                \end{scope}
                \addplot[
                    CasalBrand,
                    line width=1.2pt,
                ]
                table[
                    col sep=comma,
                    x=time_s,
                    y=amplitude,
                ]{data/phase_wrap_time_compressed.csv};
                \draw[
                    SiennaBrand,
                    densely dashed,
                    line width=0.8pt
                ]
                    (axis cs:0.2247,-0.23)
                    --
                    (axis cs:0.2247,0.1603744581);
                \draw[
                    SiennaBrand,
                    line width=1.0pt
                ]
                    (axis cs:0,-0.23)
                    --
                    (axis cs:0.2247,-0.23);
                \draw[
                    SiennaBrand,
                    line width=1.0pt,
                    -{Triangle[length=2.2mm,width=1.7mm]}
                ]
                    (axis cs:0.08,-0.23)
                    --
                    (axis cs:0.11235,-0.23);
                \node[
                    SiennaBrand,
                    anchor=south,
                    yshift=1.5mm
                ]
                    at (axis cs:0.11235,-0.23)
                    {$t$};
                \addplot[
                    only marks,
                    mark=*,
                    mark size=2.5pt,
                    SiennaBrand,
                ]
                coordinates {(0.2247,0.1603744581)};
            \end{axis}
            \coordinate (time-plot-east) at (current bounding box.east);
        \end{tikzpicture}
    \end{minipage}
    \hfill
    \begin{minipage}[c]{0.45\textwidth}
        \makebox[\dimexpr\linewidth+3mm\relax][r]{%
        \raisebox{1.155cm}[6cm][0cm]{%
        \begin{tikzpicture}[remember picture]
            \pgfplotstableread[
                col sep=comma
            ]{data/phase_wrap_time_compressed.csv}\phasewrapdata
            \def\phaseRotation{0}
            \def\phaseViewAzimuth{-60}
            \begin{axis}[
                width=1.45\linewidth,
                height=6.0cm,
                view={\phaseViewAzimuth}{8},
                axis lines=none,
                xmin=-1.2,
                xmax=1.2,
                ymin=-1.2,
                ymax=1.2,
                zmin=-0.25,
                zmax=1.15,
                z post scale=1.9,
            ]
                \foreach \row in {180,190,...,1370}{
                    \pgfmathtruncatemacro{\nextrow}{\row+10}
                    \pgfplotstablegetelem{\row}{time_s}\of\phasewrapdata
                    \pgfmathsetmacro{\curtaint}{\pgfplotsretval}
                    \pgfplotstablegetelem{\row}{amplitude}\of\phasewrapdata
                    \pgfmathsetmacro{\curtainz}{
                        \pgfplotsretval+0.16
                    }
                    \pgfmathsetmacro{\curtaintheta}{
                        360*\curtaint/0.84+\phaseRotation
                    }
                    \pgfmathsetmacro{\curtainx}{
                        cos(\curtaintheta)
                    }
                    \pgfmathsetmacro{\curtainy}{
                        sin(\curtaintheta)
                    }
                    \pgfplotstablegetelem{\nextrow}{time_s}\of\phasewrapdata
                    \pgfmathsetmacro{\curtaintnext}{\pgfplotsretval}
                    \pgfplotstablegetelem{\nextrow}{amplitude}\of\phasewrapdata
                    \pgfmathsetmacro{\curtainznext}{
                        \pgfplotsretval+0.16
                    }
                    \pgfmathsetmacro{\curtainthetanext}{
                        360*\curtaintnext/0.84+\phaseRotation
                    }
                    \pgfmathsetmacro{\curtainxnext}{
                        cos(\curtainthetanext)
                    }
                    \pgfmathsetmacro{\curtainynext}{
                        sin(\curtainthetanext)
                    }
                    \pgfmathsetmacro{\curtainmidtheta}{
                        0.5*(\curtaintheta+\curtainthetanext)
                    }
                    \pgfmathparse{
                        sin(\curtainmidtheta-\phaseViewAzimuth) > 0
                        ? 0.08
                        : 0.15
                    }
                    \edef\shadeopacity{\pgfmathresult}
                    \edef\curtainpanel{
                        \noexpand\path[
                            fill=CasalBrand,
                            fill opacity=\shadeopacity,
                            draw=none
                        ]
                            (axis cs:\curtainx,\curtainy,-0.02)
                            --
                            (axis cs:\curtainx,\curtainy,\curtainz)
                            --
                            (
                                axis cs:\curtainxnext,
                                \curtainynext,
                                \curtainznext
                            )
                            --
                            (
                                axis cs:\curtainxnext,
                                \curtainynext,
                                -0.02
                            )
                            -- cycle;
                    }
                    \curtainpanel
                }

                \pgfmathsetmacro{\peaktheta}{
                    360*0.8393/0.84+\phaseRotation
                }
                \pgfmathsetmacro{\peakx}{cos(\peaktheta)}
                \pgfmathsetmacro{\peaky}{sin(\peaktheta)}
                \pgfmathsetmacro{\peakz}{0.45+0.16}
                \edef\peakline{
                    \noexpand\draw[
                        gray!70,
                        densely dashed,
                        line width=0.8pt
                    ]
                        (axis cs:\peakx,\peaky,-0.02)
                        --
                        (axis cs:\peakx,\peaky,\peakz);
                }
                \peakline

                \addplot3[
                    black,
                    line width=1.0pt,
                    domain=0:360,
                    samples=100,
                ] ({cos(x)},{sin(x)},{-0.02});

                \addplot3[
                    CasalBrand,
                    line width=1.2pt,
                    restrict expr to domain={
                        \thisrow{time_s}
                    }{0:0.839999},
                    unbounded coords=jump,
                ]
                table[
                    col sep=comma,
                    x expr={
                        cos(
                            deg(2*pi*\thisrow{time_s}/0.84)
                            +\phaseRotation
                        )
                    },
                    y expr={
                        sin(
                            deg(2*pi*\thisrow{time_s}/0.84)
                            +\phaseRotation
                        )
                    },
                    z expr={\thisrow{amplitude}+0.16},
                ]{data/phase_wrap_time_compressed.csv};

                \pgfmathsetmacro{\secondarytheta}{
                    360*0.2247/0.84+\phaseRotation
                }
                \pgfmathsetmacro{\secondaryx}{cos(\secondarytheta)}
                \pgfmathsetmacro{\secondaryy}{sin(\secondarytheta)}
                \pgfmathsetmacro{\secondaryz}{0.1603744581+0.16}
                \pgfmathsetmacro{\sectorstart}{\peaktheta-360}
                \path[
                    fill=SiennaBrand,
                    fill opacity=0.25,
                    draw=SiennaBrand,
                    line width=1.0pt
                ]
                    (axis cs:0,0,-0.02)
                    --
                    plot[
                        domain=\sectorstart:\secondarytheta,
                        samples=30
                    ]
                    (
                        axis cs:{cos(\x)},
                        {sin(\x)},
                        -0.02
                    )
                    -- cycle;
                \pgfmathsetmacro{\sectormid}{
                    0.5*(\sectorstart+\secondarytheta)
                }
                \pgfmathsetmacro{\sectorarrowstart}{\sectormid-12}
                \addplot3[
                    SiennaBrand,
                    line width=1.0pt,
                    -{Triangle[length=3mm,width=2.2mm]},
                    domain=\sectorarrowstart:\sectormid,
                    samples=8,
                ]
                ({cos(x)},{sin(x)},{-0.02});
                \pgfmathsetmacro{\sectormidx}{cos(\sectormid)}
                \pgfmathsetmacro{\sectormidy}{sin(\sectormid)}
                \edef\sectorlabel{
                    \noexpand\node[
                        SiennaBrand,
                        font=\noexpand\large,
                        anchor=south,
                        yshift=-0.5mm
                    ]
                        at (
                            axis cs:\sectormidx,
                            \sectormidy,
                            -0.02
                        )
                        {$\noexpand\phi$};
                }
                \sectorlabel
                \edef\phaseradii{
                    \noexpand\draw[
                        SiennaBrand,
                        line width=0.8pt
                    ]
                        (axis cs:0,0,-0.02)
                        --
                        (axis cs:\peakx,\peaky,-0.02);
                    \noexpand\draw[
                        SiennaBrand,
                        line width=0.8pt
                    ]
                        (axis cs:0,0,-0.02)
                        --
                        (
                            axis cs:\secondaryx,
                            \secondaryy,
                            -0.02
                        );
                }
                \phaseradii
                \edef\secondaryline{
                    \noexpand\draw[
                        SiennaBrand,
                        densely dashed,
                        line width=0.8pt
                    ]
                        (
                            axis cs:\secondaryx,
                            \secondaryy,
                            -0.02
                        )
                        --
                        (
                            axis cs:\secondaryx,
                            \secondaryy,
                            \secondaryz
                        );
                }
                \secondaryline
                \addplot3[
                    only marks,
                    mark=*,
                    mark size=2.5pt,
                    SiennaBrand,
                ]
                coordinates {
                    (\secondaryx,\secondaryy,\secondaryz)
                };

                \addplot3[
                    black,
                    line width=1.0pt,
                    -{Triangle[length=3mm,width=2.2mm]},
                    domain=180:210,
                    samples=12,
                ] ({cos(x)},{sin(x)},{-0.02});
            \end{axis}
            \coordinate (phase-plot-west) at (current bounding box.west);
            \path[overlay]
                (current bounding box.south |- time-label.base)
                node[
                    name=phase-label,
                    anchor=base,
                    font=\large
                ]
                {Phase, $\phi\in\mathbb{S}^1$};
        \end{tikzpicture}
        }
        }
    \end{minipage}
    \begin{tikzpicture}[remember picture,overlay]
        \draw[
            /tikz/commutative diagrams/every arrow,
            line width=0.8pt,
            black,
            shorten <=3pt,
            shorten >=3pt
        ]
            (time-plot-east)
            --
            node[midway, above=1mm, font=\large]
                {$\Phi\colon\mathbb{R}\rightarrow\mathbb{S}^1$}
            (phase-plot-west |- time-plot-east);
    \end{tikzpicture}
\end{center}

\begingroup
\setlength{\abovecaptionskip}{0pt}
\nonfloatcaption{\textbf{Phase reparameterisation of the cardiac cycle.} A heart beat between consecutive R peaks is mapped from linear time $t\in\mathbb{R}$ to phase $\phi\in\mathbb{S}^1$. Identifying $\phi=0$ with $\phi=2\pi$ makes the periodic structure of the heartbeat explicit while preserving within-cycle ordering.}
\label{fig:phase-wrap}
\endgroup
\vspace{1em}

%% file: 01_introduction.tex
\section{Introduction}

The 12-lead electrocardiogram (ECG) provides a rapid, non-invasive measurement of cardiac electrical activity at millisecond-scale resolution. Its interpretation depends on the morphology and timing of the P wave, QRS complex, T wave, and intervening intervals \citep{goldberger2024clinical}. Despite variation in rate and morphology, cardiac electrical activity follows a recurrent sequence of depolarisation and repolarisation \citep{guyton2006textbook}. Mapping successive R--R intervals onto a common phase coordinate preserves this ordering while normalising beat duration, a principle reflected in QT rate correction \citep{plagwitz2026rlign}. Conduction delay and ischaemic deformation can then be viewed as changes to within-cycle timing and morphology, while ectopic activity disrupts the recurring pattern across cycles.

Continuous physiological recordings capture both patient-state dynamics and sensor-level noise---motion, electrode contact, baseline wander---in a single observed trace. Scored in input space, autoregressive and reconstruction objectives are compelled to reward reproducing the \emph{sensor} observation in full, noise included. Joint-embedding predictive architectures (JEPAs) offer an alternative prescription, whereby predictions are scored against a latent target, training an encoder--predictor system to anticipate a future or masked representation of the signal state rather than the raw sample \citep{lecun2022path,assran2023ijepa}. Under this paradigm, sensor variation need not be reproduced sample by sample; insofar as the target encoder suppresses variation unnecessary for predicting the latent signal state, representational capacity can concentrate on physiological dynamics.

Learning the heart’s recurring structure requires exactly this kind of objective: one that preserves cardiac-cycle timing and morphology without being dominated by nuisance signals. Pointwise reconstruction can overemphasise high-amplitude motion and channel artefacts; the corresponding advantage of latent prediction has been reported for ECG \citep{kim2024ecgjepa} and EEG \citep{panchavati2026laya}. The relaxation does not, however, confer invariance to artefacts: it withdraws the explicit reward for reproducing them without penalising their encoding. Standard latent-prediction objectives also leave the organisation of cardiac phase unspecified. A model may therefore encode phase without organising latent motion consistently across beats.

In this work, we introduce \Winder{}, a phase-equivariant extension of LeJEPA \citep{balestriero2025lejepa,maes2026leworldmodel}. \Winder{} retains LeJEPA’s regularisation against representational collapse and adds a transport objective that penalises deviations from phase equivariance. This objective encourages latent trajectories to follow cardiac-phase order consistently across beats, grounding the representation in the heart’s physiological cycle.

Empirically, enforcing this periodic symmetry improves diagnostic performance over a matched transport-free control. \Winder{} also attains performance in the same broad range as substantially larger self-supervised ECG models with only $1.2\,\mathrm{M}$ deployed parameters, while producing phase-ordered latent geometry.

The remainder of this paper is organised as follows. Next, the current state of the art of self-supervised learning, time-series representation learning, and equivariant deep learning is discussed. Inspired by these ideas, \Cref{sec:theory} derives the required loss function to enforce phase-equivariant representations. In \Cref{sec:method}, the implementation of \Winder{} in a compact causal architecture is described. Our experiments in \Cref{sec:results} assess diagnostic performance, cardiac-phase equivariance, latent geometry, and the contribution of the transport objective. Finally, the implications and limitations of these findings are discussed in \Cref{sec:discussion} before concluding in \Cref{sec:conclusions}.

%% file: 02_related_work.tex

\section{Related Work}
\label{sec:related-work}

\Winder{} sits at the intersection of self-supervised representation learning and geometric deep learning, using a phase-aware equivariant latent space to capture the cyclic structure of the cardiac cycle in ECG time series.
Accordingly, this work draws on three active fields: predictive
self-supervised learning, time-series representation learning at constrained model scale, and equivariant (symmetry-structured) representation learning.

\subsection{Predictive Self-Supervised Learning}

A central challenge in self-supervised learning is preventing non-contrastive objectives from converging to uninformative representations.
Recently, LeJEPA \citep{balestriero2025lejepa} prevented representation collapse by regularising the embedding distribution directly, rather than through heuristics such as stop-gradients or teacher--student networks \citep{grill2020byol}.

LeJEPA pairs a predictive loss with the Sketched Isotropic Gaussian Regulariser (SIGReg), which tests Gaussian isotropy of the embedding distribution along randomly sampled projection directions
\begin{equation}
\begin{aligned}
\mathrm{SIGReg}_{\mathcal T}(z)
    &\coloneqq \frac{1}{M}\sum_{m=1}^{M}\mathcal T(a_m;z), \\
a_m &\overset{\mathrm{iid}}{\sim}
    \operatorname{Unif}(\mathbb S^{K-1}),
\end{aligned}
\label{eq:sigreg}
\end{equation}
where $\mathcal T(a;z)$ is a univariate goodness-of-fit statistic applied
to the projection of the latent embeddings, $z \in \mathbb R^K$, along the unit vector, $a$.
To target an isotropic Gaussian distribution, the natural univariate test is the Epps--Pulley (EP) goodness-of-fit statistic \citep{epps1983test}:
%
\begin{equation}
\mathrm{EP}(a;z) \coloneqq N\int_{\mathbb R}
    \left|\widehat{\varphi}_a(\tau)-\varphi_0(\tau)\right|^2
    \varphi_0(\tau)\,d\tau.
\label{eq:epps-pulley}
\end{equation}
The empirical and standard Gaussian characteristic functions are defined as
\begin{equation}
\widehat{\varphi}_a(\tau)
    \coloneqq \frac1N\sum_{n=1}^N e^{i\tau\,a^\top z_n},
    \quad
\varphi_0(\tau) \coloneqq e^{-\tau^2/2}.
\label{eq:characteristic-functions}
\end{equation}
Equation~\eqref{eq:sigreg} replaces the maximum over projection directions in LeJEPA's theoretical criterion with an average, preventing the gradient from concentrating on only the worst-performing direction. Collapse is thus excluded by construction of the SIGReg regulariser rather
than by tuning, leaving only a single trade-off hyperparameter that tunes the relative weight between regulariser and predictive loss. 
LeWorldModel \citep{maes2026leworldmodel} carries this recipe to the temporal setting: next-embedding prediction from pixels under the same SIGReg regularisation.

Since LeJEPA, the recipe's every component has been revised in turn: the target geometry, rectified toward sparsity \citep{kuang2026lpjepa}, or proven suboptimal to a uniform sphere on manifolds \citep{nicollier2026spherejepa}; 
the test statistic \citep{wu2026visreg}; and the constraint's ambient scope, which Sub-JEPA restricts to random, deliberately frozen subspaces \citep{zhao2026subjepa}.

A parallel wave targets the marginal's blindness to temporal order: SD-JEPA learns a progression coordinate \citep{thil2026sdjepa}; letting one emerge from temporally centered residuals \citep{liu2026tcsigreg}; or pools the statistic within-sample across time \citep{chemeris2026lenepa}. Yet to our knowledge the regularised objective remains a set in every case, invariant to permuting the time index.
The conditions under which the base recipe's guarantees hold are only now being characterized, so far confined to a restricted world class of independent, Ornstein--Uhlenbeck latent dynamics under isotropic sampling \citep{klindt2026lejepaworldmodel}.

SIGReg has recently entered biosignal representation learning, primarily through EEG and broader multi-domain time-series models \citep{broustail2026lumamba,panchavati2026laya,petersen2026hepa}. Its application to ECG remains limited. LeNEPA regularises temporal embeddings within individual PTB-XL recordings \citep{chemeris2026lenepa}, while Beyond Patient Invariance represents pathology as a transition in latent space and reports that SIGReg outperforms VICReg on physiological signals \citep{fernandes2026patientinvariance}. Neither method imposes a cardiac phase coordinate.

CardioState-JEPA is the closest phase-aware ECG method, but treats phase as an auxiliary supervised target within an objective that retains an EMA teacher \citep{shafiq2026cardiostatejepa}. Sonata models another periodic biosignal—gait and tremor dynamics—by expanding the recurrence eigenspectrum to support oscillatory modes \citep{delaney2026sonata,grazzi2025negativeeigenvalues}. In both cases, cyclicity is predicted or accommodated rather than imposed as a symmetry of the latent space.

Other ECG self-supervised methods either predate SIGReg \citep{kim2024ecgjepa} or use world models for intervention simulation \citep{chen2026ecgwm}. Moreover, a recent scaling study finds contrastive predictive coding modestly ahead of JEPA-style objectives at comparable scale \citep{almasud2026ecgscaling}. Thus, neither the appropriate anti-collapse regulariser nor the correct representation of cardiac phase is settled. \Cref{app:ptbxl} places \Winder{} alongside these methods on PTB-XL, with their parameter counts and pre-training corpora, and sets out why no two published entries share an evaluation protocol.

\subsection{Time-Series Foundation Models: Scale, Adaptation, and Efficiency}

Outside this lineage, general-purpose time-series foundation models continue to diversify along three largely independent axes \citep{laglil2026tsfmsurvey}: scale, in-context adaptation, and architectural efficiency. 
Pretraining now extends into wearable sensing at trillion-minute scale, with SensorFM's own scaling ablations spanning up to a hundred million parameters and transferring across dozens of downstream health tasks \citep{narayanswamy2026sensorfm}. 
A parallel direction replaces retraining with adaptation at inference time:
Baguan-TS conditions a sequence-native forecaster on a retrieved,
covariate-aware context rather than updating any weight
\citep{yang2026baguants}. 
At the other end of the spectrum, with only 0.2M--2.6M parameters, Reverso \citep{fu2026reverso} closes much of the performance gap to models one to three orders of magnitude larger. It does this through a hybrid architecture alternating gated long convolutions with DeltaNet's linear-recurrent state updates. \Winder{} sits at a similar scale to Reversoby imposing structure on the latent space rather than architecturaly innovations.

Two related time-series architectures, Olivia and RoMAE, also apply rotations to continuous coordinates, but toward different ends than \Winder{}. Olivia harmonises cross-domain spectral heterogeneity by learning an
orthogonal transform of the raw time axis, parameterised, as here
(\cref{eq:householder-basis}), as a product of Householder reflections
\citep{fei2026olivia}; RoMAE instead fixes the rotation, applying rotary
position embeddings to continuous positions within a masked-autoencoder
objective \citep{zivanovic2025romae}. In both cases the rotated coordinate is domain-agnostic --- dataset identity or sequence position --- and the rotation removes or encodes a nuisance; \Winder{}'s rotation instead acts on a single physiologically derived coordinate, cardiac phase, to impose a symmetry the latent representation is required to respect.

\subsection{Equivariant Representation Learning for Time Series}
%
%

Equivariant networks encode a known symmetry as a group action the model is constrained to respect: transforming the input by $g$ transforms the representation by a corresponding $R_g$. The canonical construction is group-equivariant convolution \citep{cohen2016gcnn}; Bronstein et al.\ \citep{bronstein2021gdl} place this construction in the geometric deep learning catalogue of grids, groups, graphs, geodesics, and gauges.
Two symmetries spring to mind for time series: time-translation (sliding the window) and dilation (changing sampling rate). Both have been realised as group-equivariant layers \citep{romero2023wavelet,worrall2019dss}, including sampling-rate and discrete-scale forms \citep{graf2025flowstate,morandi2026sewavenet}. A time shift and a phase advance are not, however, the symmetry that defines cardiac cycle: phase is periodic in the beat, not a shift or a stretch of the recording.

The constructions above constrain the geometry of the input: the layer is built so that a transformation of $x$ is matched by a transformation of the features. A second literature constrains the geometry of the latent space instead, requiring $z$ itself to transform under $G$. A disentangled representation is defined as a split of $z$ into an invariant factor and a factor that transforms with the group \citep{higgins2018disentangled}. Unsupervised group-equivariant autoencoders learn that split from a known action on the input, producing invariant and equivariant latent coordinates \citep{winter2022unsupervised}. Class--pose decomposition writes $z$ as an invariant class code times a group element \citep{marchetti2022classpose}. The same move appears outside vision: POEM rotates paired latent features of a time-series forecaster by a learned $SO(2)$ phase under periodicity drift \citep{zhu2026poem}, and PHALAR realises time-shift as a $U(1)$ rotation via the Fourier shift theorem, in audio \citep{marincione2026phalar}. These methods show that a group action can be imposed on $z$, but they are developed for images, objects, generic time series, and audio, not for healthcare.

On ECG, physically grounded latent geometry exists without a group action. LVCG maps 12-lead recordings to a latent Frank vectorcardiogram and trains view-invariant tokens; R-peaks segment beats, but they are not used as a symmetry \citep{huang2026lvcg}. Nef-Net queries an implicit electrocardiac field by viewpoint angle to synthesise new leads \citep{chen2021nefnet}. In both cases the geometry is of the source as seen from different electrodes. Invariance to view is not equivariance to cardiac phase: transforming the input should move $z$ in a controlled way, not leave it fixed.

Similar approaches appear in health, but none have imposed phase equivariance on a learned ECG representation. CatEquiv uses circular 1-D convolution so that rolling an inertial window is equivariant by construction \citep{maruyama2025catequiv}: that is a cyclic shift of the recording, not a phase advance through a physiological cycle. Echo-POSED is equivariant to probe pose and invariant to cardiac phase, so the representation does not track left-ventricular twist \citep{stenhede2026echoposed}. This phase invariance inverts the constraint required here.

To our knowledge, a derived cardiac-phase action on a learned ECG latent remains unoccupied. This work fills this gap.

%% file: 03_theory.tex

\definecolor{SiennaBrand}{HTML}{EB684C}
\definecolor{CasalBrand}{HTML}{35636E}

\section{Theory}
\label{sec:theory}
The recurring physiological organisation of the cardiac cycle motivates treating phase not merely as a timestamp, but as a symmetry that the latent representation should preserve. Given this observation, instead of parameterising the heartbeat as a function of time, $t \in \mathbb{R}$, it can be reparameterised as a function of phase, $\phi \in [0, 2\pi) \cong \mathbb{S}^1$, representing the closed-loop cardiac cycle.

Given detected R-peak timestamps $\{R_i\}$, the phase map $\Phi:\mathbb{R}\to\mathbb{S}^1$ is defined by
\begin{equation}
    t \longmapsto
    \phi(t)
    \coloneqq
    2\pi\frac{t-R_i}{R_{i+1}-R_i},
    \quad t\in[R_i,R_{i+1}).
    \label{eq:phase-map}
\end{equation}
This reparameterisation introduces the phase-translation group $\mathcal{T} \coloneqq \{T_\Delta \mid \Delta \in [0, 2\pi)\}$, which is isomorphic to $U(1)$.
Let $\mathcal{X}$ denote a space of phase-parameterised multivariate time series on which $\mathcal{T}$ acts --- in this work a 12-lead ECG signal. For $T_\Delta\in\mathcal{T}$, the action on $x\in\mathcal{X}$ is defined by

\begin{equation}
    (T_\Delta x)(\phi) \coloneqq x\bigl((\phi - \Delta) \bmod 2\pi\bigr).
    \label{eq:phase-shift}
\end{equation}

This definition corresponds to the canonical left action of $U(1)$ shifting functions along the circular domain. The central idea of \Winder{} is simple: the latent space should obey the same symmetry as the cardiac cycle. The following derivation establishes the equivariance condition required for the latent representation to satisfy the periodic symmetry.


\subsection{Phase-Equivariant Latent Space}
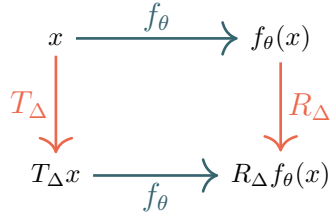
\begin{figure}[!t]
    \centering
    \begin{tikzcd}[
        column sep=4.75em,
        row sep=3.4em,
        every node/.append style={font=\large},
        every label/.append style={font=\large}
    ]
        x
        \arrow[r, "\textcolor{CasalBrand}{f_\theta}", draw=CasalBrand, line width=1pt]
        \arrow[d, "\textcolor{SiennaBrand}{T_\Delta}"', draw=SiennaBrand, line width=1pt]
        & f_\theta(x)
        \arrow[d, "\textcolor{SiennaBrand}{R_\Delta}", draw=SiennaBrand, line width=1pt] \\
        T_\Delta x
        \arrow[r, "\textcolor{CasalBrand}{f_\theta}"', draw=CasalBrand, line width=1pt]
        & R_\Delta f_\theta(x)
    \end{tikzcd}
    \caption{\textbf{Commutative diagram of phase equivariance.} Encoding after a
    phase translation $T_\Delta$ is equivalent to applying the
    corresponding latent transformation $R_\Delta$ after encoding,
    $f_\theta(T_\Delta x)=R_\Delta f_\theta(x)$.}
    \label{fig:encoder-equivariance}
\end{figure}

\input{latent_space_rotation}

To model latent evolution over cardiac phase, the encoder is required to preserve the phase-translation symmetry of the input. Specifically, the action $T_\Delta$ on the input space $\mathcal{X}$ should induce a corresponding action $R_\Delta$ on the latent space $\mathcal{Z}$, such that
\begin{equation}
    f_\theta(T_\Delta x) = R_\Delta f_\theta(x),
    \qquad \Delta \in [0, 2\pi).
    \label{eq:latent-equivariance}
\end{equation}
Thus, translating a signal in phase before encoding is equivalent to translating its latent representation after encoding (see \Cref{fig:encoder-equivariance}).

The natural building blocks of the latent group symmetry are the irreducible representations (irreps) of the underlying group, $U(1)$.
For convenience, the analysis uses real representations.

Under the isomorphism $U(1) \cong SO(2)$, each conjugate pair of non-trivial complex characters is represented by a two-dimensional real rotation block. The real irreducible representations are of two types:
\begin{equation}
    \rho_\omega(\Delta) =
    \begin{cases}
        1, & \omega = 0, \\[4pt]
        \begin{pmatrix}
            \cos(\omega\Delta) & -\sin(\omega\Delta) \\
            \sin(\omega\Delta) & \cos(\omega\Delta)
        \end{pmatrix},
        & \omega > 0.
    \end{cases}
    \label{eq:rotation-irrep}
\end{equation}
Imposing $2\pi$-periodicity, $R_{2\pi}\overset{!}{=}\mathbf{I}_K$, gives
\begin{equation}
    \chi_\omega(2\pi)=1
    \quad\Longleftrightarrow\quad
    \omega=n,\quad n\in\mathbb{Z}.
    \label{eq:frequency-quantisation}
\end{equation}
The zero-frequency character is trivial. Over $\mathbb{R}$, the characters
$n$ and $-n$ form equivalent two-dimensional rotation representations, so
only the canonical representatives $n\in\mathbb{Z}_{>0}$ are retained.

Because $U(1)$ is compact, averaging an inner product over the group yields an invariant inner product. Consequently, every finite-dimensional continuous real representation is equivalent, after a change of basis, to an orthogonal representation.
Because the encoder's latent coordinates are learned, their axes need not align with the invariant subspaces on which the irreducible representations act. An orthogonal change of basis $Q\in O(K)$ is therefore introduced, with $Q^\top z$ expressing the learned latent in the canonical irrep basis.

In this canonical basis, the group action decomposes into a trivial component and harmonic isotypic components, so the latent space admits the direct-sum decomposition
\begin{equation}
    \mathbb{R}^K
    \cong
    \mathbb{R}^{K_0}
    \oplus
    \bigoplus_{n=1}^{n_{\max}}
    \left(\mathbb{R}^{c_n}\otimes\mathbb{R}^2\right).
\end{equation}
Accordingly, the transport action is block diagonal in the coordinates $Q^\top z$:
\tcbset{
    highlight math style={
        colback=SiennaBrand!10,
        colframe=SiennaBrand,
        boxrule=1pt,
        arc=2mm,
        left=2pt,
        right=2pt,
        boxsep=0pt
    }
}
\begin{subequations}
    \label{eq:canonical-block-form}
    \begin{empheq}[box=\tcbhighmath]{align}
        R_\Delta &= Q B_\Delta Q^\top,
        \label{eq:canonical-transport} \\
        B_\Delta
        &= \mathbf{I}_{K_0}
        \oplus
        \bigoplus_{n=1}^{n_{\max}}
        \left(\mathbf{I}_{c_n}\otimes\rho(n\Delta)\right).
        \label{eq:canonical-block-decomposition}
    \end{empheq}
\end{subequations}
Here, $c_n\in\mathbb{Z}_{\geq 0}$ is the multiplicity of harmonic $n$: the number of independent two-dimensional feature channels that transform at that rate. The latent dimension is therefore
\begin{equation}
    K
    =
    \underbrace{K_0}_{\text{invariant capacity}}
    +
    \underbrace{2\sum_{n=1}^{n_{\max}}c_n}_{\text{harmonic capacity}}.
    \label{eq:latent-capacity}
\end{equation}
The harmonic orders $n$ determine which phase-dependent transformations the latent space can represent, whereas the multiplicities $c_n$ determine how much feature capacity is allocated to each order.
\Cref{fig:canonical-block-transport} makes the transformation of an individual latent vector explicit. In the canonical coordinates $Q^\top z$, the $K_0$ invariant coordinates remain fixed, while each of the $c_n$ two-dimensional pairs assigned to harmonic $n$ rotates by $n\Delta$, so different components of the same latent vector advance at different integer multiples of cardiac phase.

\begin{figure}[t]
    \centering
    \begingroup
    \delimiterfactor=1250
    \delimitershortfall=-2pt
    \resizebox{\linewidth}{!}{$\displaystyle
        \begin{pmatrix}
            \rule{0pt}{2.75ex}
            \textcolor{SiennaBrand}{\mathbf{I}_{K_0}}
                & 0 & 0 & \cdots & 0 \\
            0 & \textcolor{CasalBrand}{\mathbf{I}_{c_1}\otimes\rho(\Delta)}
                & 0 & \cdots & 0 \\
            0 & 0 & \textcolor{CasalBrand}{\mathbf{I}_{c_2}\otimes\rho(2\Delta)}
                & \ddots & \vdots \\
            \vdots & \vdots & \ddots & \ddots & 0 \\
            0 & 0 & \cdots & 0
                & \textcolor{CasalBrand}{\mathbf{I}_{c_{n_{\max}}}\otimes\rho(n_{\max}\Delta)}
            \rule[-1.15ex]{0pt}{0pt}
        \end{pmatrix}
    $}
    \endgroup
    \caption{\textbf{Canonical block structure of the transport operator
    $B_\Delta$.} The $\mathbf{I}_{K_0}$ block
    (\textcolor{SiennaBrand}{orange}) is fixed under transport; the harmonic
    blocks (\textcolor{CasalBrand}{teal}) rotate by $n\Delta$ via
    $\rho(n\Delta)$, each acting on $c_n$ independent two-dimensional
    channels.}
    \label{fig:canonical-block-transport}
\end{figure}

More generally, the orthogonal basis $Q$ may also be learned. One
parameterisation uses a product of $J$ Householder reflections,
\begin{equation}
    Q=\prod_{j=1}^{J}H(v_j),
    \qquad
    H(v)=I-2\frac{vv^\top}{v^\top v},
    \label{eq:householder-basis}
\end{equation}
which guarantees $Q^\top Q=I$ throughout training
\citep{mhammedi2017efficient}. The selected architecture in
\cref{sec:method} instead fixes the canonical basis, so its transport operator
has no learnable parameters.

\subsection{Learning Equivariant Transport}

The \Winder{} objective combines latent prediction, phase-equivariant
transport, and spectral regularisation:
\begin{equation}
    \mathcal{L}_{\mathrm{Winder}}
    =
    {\color{CasalBrand}
        \mathcal{L}_{\mathrm{pred}}
        + \lambda_{\mathrm{SIG}}\mathcal{L}_{\mathrm{SIG}}
    }
    + {\color{SiennaBrand}
        \lambda_{\mathrm{trans}}\mathcal{L}_{\mathrm{trans}}
    }.
    \label{eq:winder-total-loss}
\end{equation}
Here, the \textcolor{CasalBrand}{blue} terms are inherited from LeJEPA:
$\mathcal{L}_{\mathrm{pred}}$ trains the latent predictor by requiring the
latent magnitude and content to support next-token prediction, while
$\mathcal{L}_{\mathrm{SIG}}$ prevents dimensional collapse by controlling the
full latent distribution. The \textcolor{SiennaBrand}{orange} term,
$\mathcal{L}_{\mathrm{trans}}$, is specific to \Winder{} and enforces phase
equivariance by constraining how the unit-normalised latent direction changes
with cardiac phase.

\subsubsection{Latent Prediction}
\label{sec:latent-prediction}


Given a context ending at token $t$, the causal predictor $g_\psi$ maps the
projected latent prefix $z_{\leq t}$ to a prediction of the next latent,
\begin{equation}
    \hat{z}_{t+1}
    \coloneqq
    g_\psi\!\left(z_{\leq t}\right)_{t+1}.
\end{equation}
The prediction loss is the mean-squared error between the predictor output $\hat{z}$ and the projection head's own output $z$, averaged over latent dimensions and target positions.

\subsubsection{Spectral Regularisation}

\begin{table*}[t]
    \centering
    \small
    \begin{tabular}{llll}
        \toprule
        ECG structure & Physiological process & Typical duration [ms] & Characteristic order \\
        \midrule
        P wave        & Atrial depolarisation                 & 80--110    & $n=10$--$13$ \\
        PR interval   & Atrial-to-ventricular conduction      & 120--200   & $n=5$--$9$ \\
        QRS complex   & Ventricular depolarisation            & 70--110    & $n=10$--$15$ \\
        ST segment    & Ventricular plateau phase             & 80--120    & $n=9$--$13$ \\
        T wave        & Ventricular repolarisation            & 100--250   & $n=4$--$10$ \\
        QT interval   & Total ventricular electrical activity & 350--450   & $n=3$ \\
        R--R interval & Complete cardiac cycle                & 600--1000  & $n=1$ \\
        \bottomrule
    \end{tabular}
    \caption{\textbf{Characteristic timescales of ECG structure.}
    Physiological durations are representative adult values
    \citep{guyton2006textbook}. Characteristic orders follow
    \cref{eq:harmonic-resolution}, using $T=1000$ ms;
    sharper morphology may require higher orders.}
    \label{tab:ecg-timescales}
\end{table*}
\label{sec:transport}

The prediction and transport objectives do not prevent the encoder from collapsing onto a low-dimensional representation. SIGReg therefore constrains the latent distribution towards a reference distribution $p_0$.

SIGReg follows LeJEPA's standard random-projection prescription. Rather than comparing distributions directly in the $K$-dimensional latent space, SIGReg compares their one-dimensional sketches along random unit directions $a_m\in\mathbb{S}^{K-1}$.
Matching random one-dimensional projections provides a scalable surrogate for matching the full latent distribution. In practice, the integral is evaluated by numerical quadrature and the projection directions are resampled throughout training.

Once a distribution can be targeted, the remaining question is which distribution to use. To avoid competing with the transport objective, the reference distribution must be invariant under the latent group action. A target is called admissible if
\begin{equation}
    (R_\Delta)_{\#}p_0=p_0,
    \quad \forall \Delta,
    \label{eq:admissible-target}
\end{equation}
where $(R_\Delta)_{\#}p_0$ denotes the pushforward of $p_0$ under $R_\Delta$, namely the distribution obtained by applying $R_\Delta$ to samples from $p_0$.

For a centred Gaussian target, $p_0=\mathcal{N}(0,\Sigma)$, a linear transformation gives
\[
    z\sim\mathcal{N}(0,\Sigma)
    \quad\Longrightarrow\quad
    R_\Delta z\sim
    \mathcal{N}\!\left(0,R_\Delta\Sigma R_\Delta^\top\right).
\]
Admissibility therefore requires
\begin{equation}
    R_\Delta\Sigma R_\Delta^\top
    \overset{!}{=}
    \Sigma,
    \quad \forall \Delta.
    \label{eq:admissible-covariance}
\end{equation}
In the canonical basis, the most general symmetric admissible covariance is
\begin{equation}
    \begin{aligned}
        Q^\top\Sigma Q
        &=
        S_0
        \oplus
        \bigoplus_{n=1}^{n_{\max}}
        \Biggl[
            {\color{CasalBrand}
                A_n\otimes
                \begin{pmatrix}
                    1 & 0 \\
                    0 & 1
                \end{pmatrix}
            } \\
        &\qquad\qquad
            +
            {\color{SiennaBrand}
                C_n\otimes
                \begin{pmatrix}
                    0 & -1 \\
                    1 & 0
                \end{pmatrix}
            }
        \Biggr].
    \end{aligned}
\end{equation}
The \textcolor{CasalBrand}{$A_n$} term represents covariance between corresponding coordinates of different copies of harmonic $n$. The \textcolor{SiennaBrand}{$C_n$} term represents covariance between orthogonal coordinates---one copy's cosine-like coordinate and another's sine-like coordinate---corresponding to a phase offset of $\pi/2$. Here,
\[
    S_0=S_0^\top,\qquad
    A_n=A_n^\top,\qquad
    C_n=-C_n^\top,
\]
and each covariance block is required to be positive definite. When $c_n=1$, skew symmetry forces $C_n=0$, reducing the harmonic block to $\sigma_n^2\mathbf{I}_2$.

Different members of this admissible family can encode covariance structures motivated by the underlying mechanism or quantity of interest. For simplicity, however, the LeJEPA choice $\Sigma=\mathbf{I}_K$ is adopted, giving $p_0=\mathcal{N}(0,\mathbf{I}_K)$, which is invariant under every orthogonal transformation.

SIGReg consequently supplies two complementary anti-collapse pressures. First,
it penalises concentration of $z$ in a constant or low-dimensional subspace, forcing latents from different sources to occupy distinct regions of the latent space.
Second, because the isotropic target assigns variance to every latent
coordinate, it penalises concentration in the invariant block and encourages
the non-trivial harmonic irreps to remain populated. SIGReg does not determine
what those harmonic coordinates encode; the transport objective supplies the
phase-dependent structure.

\subsubsection{Phase-Equivariant Transport}


For two tokens $t$ and $t'$ from the same recording, let $\Delta_{tt'}\coloneqq(\phi(t')-\phi(t))\bmod 2\pi$ and $\hat{z}_t\coloneqq z_t/\|z_t\|_2$. Because $R_{\Delta_{tt'}}$ is orthogonal, the squared directional equivariance defect satisfies
\begin{equation}
    \begin{aligned}
        \delta_{tt'}
        &\coloneqq
        \left\|
            R_{\Delta_{tt'}}\hat{z}_t-\hat{z}_{t'}
        \right\|_2^2 \\
        &=
        2\left(
            1-
            \left\langle
                R_{\Delta_{tt'}}\hat{z}_t,\hat{z}_{t'}
            \right\rangle
        \right).
    \end{aligned}
\end{equation}
The empirical transport loss is therefore
\begin{equation}
    \mathcal{L}_{\mathrm{trans}}
    =
    \mathbb{E}_{(t,t')}
    \left[
        1-
        \left\langle
            R_{\Delta_{tt'}}\hat{z}_t,
            \hat{z}_{t'}
        \right\rangle
    \right].
\end{equation}
The transport loss enforces equivariance of unit-normalised latent directions, while latent magnitudes are controlled by the prediction and spectral objectives.

This directional objective has a phase-blind failure mode: representations can
concentrate in the invariant block, on which $R_\Delta$ acts as the identity.
Such representations may satisfy the transport constraint without using the
non-trivial harmonic irreps to encode cardiac phase. The transport objective
alone therefore does not distinguish useful phase equivariance from trivial
phase invariance.
\subsection{Bounding the Harmonic Spectrum}
\label{sec:harmonic-spectrum}

\input{gapped_harmonics_resolution}

The quantisation condition in Equation~\eqref{eq:frequency-quantisation} determines that the admissible harmonic orders are integers. The maximum retained order, $n_{\max}$, is bounded from below by the clinical timescales needed to be resolved and from above by the resolution of the sensor used to estimate phase.

Let $\delta t$ denote the shortest clinically relevant timescale within a cardiac cycle of duration $T$. Since harmonic $n$ varies on a timescale proportional to $T/n$, resolving $\delta t$ requires
\begin{equation}
    n_{\max}
    \gtrsim\frac{T}{\delta t}.
    \label{eq:harmonic-resolution}
\end{equation}
To ground this estimate physiologically, Table~\ref{tab:ecg-timescales} relates standard ECG structures to their characteristic Fourier orders. The practical upper bound on $n_{\max}$ is set by the Nyquist limit of the ECG sampling rate. At clinical ECG sampling rates the instrumental ceiling lies well above the orders in Table~\ref{tab:ecg-timescales}, so the clinical requirement is ordinarily the active constraint. This means that the number of harmonics is \emph{specified by the data} rather than found by searching over model capacity.

The highest retained harmonic and the feature capacity assigned to each
harmonic are separate design choices. For a contiguous spectrum
$n=1,\ldots,n_{\max}$ with one copy of each harmonic, every non-zero harmonic
requires a two-dimensional real rotation block. The resulting minimum latent dimension
is
\begin{equation}
    K_{\min}
    =
    K_0+2n_{\max}.
    \label{eq:minimum-latent-dimension}
\end{equation}
Additional copies, $c_n>1$, provide more independent feature channels at
harmonic $n$ without increasing the highest supported frequency. Conversely,
omitting intermediate harmonics reduces the latent dimension while preserving
the scale associated with the highest retained harmonic, but produces less
uniform localisation and stronger side lobes. \Cref{fig:gapped-harmonic-resolution}
illustrates this trade-off for the band $\{1,4,7\}$.

In practice,
$n_{\max}$ is chosen from the required physiological resolution and acquisition
rate, while $K_0$ and $\{c_n\}$ are selected on a validation partition held
out from the sealed evaluation (\cref{sec:method}).

%% file: latent_space_rotation.tex

\pgfplotstableread[col sep=comma]{data/synthetic_fourier_candidate_2.csv}{\fourierdata}

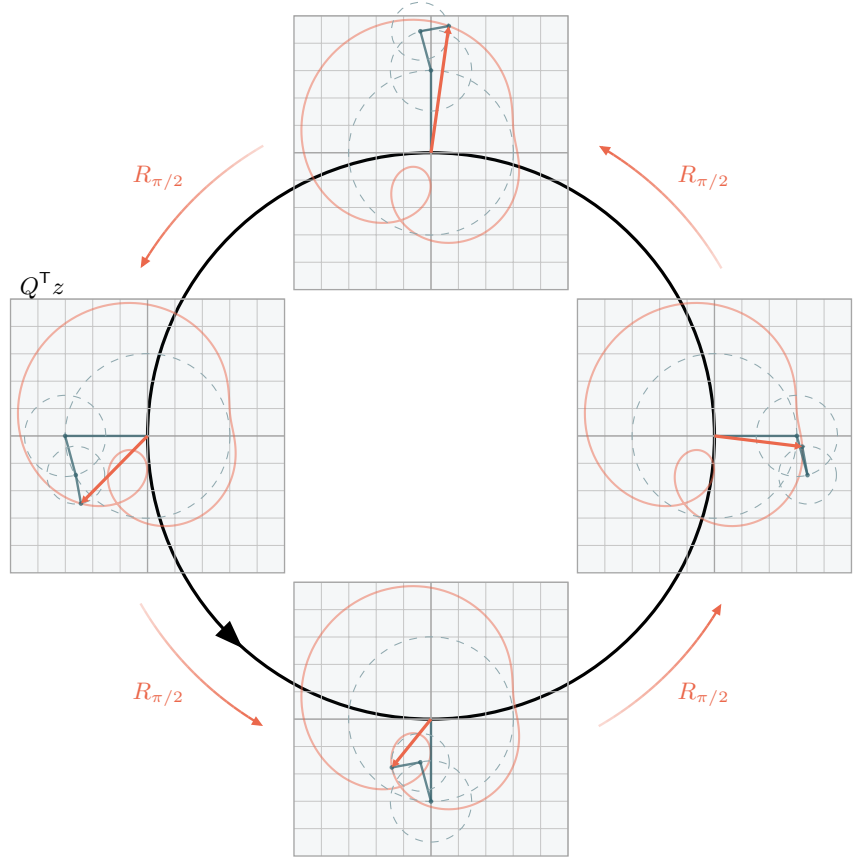
\begin{figure*}[!t]
    \centering
    \begin{tikzpicture}[scale=1.25]
        \draw[
            black,
            line width=1.2pt
        ]
            (0,0) circle[radius=3cm];
        \foreach \gridx/\gridy/\datarow/\pastrow in {
            0/3/0/90,
            -3/0/30/0,
            0/-3/60/30,
            3/0/90/60
        }{
            \begin{scope}[shift={(\gridx,\gridy)}]
                \fill[
                    CasalBrand,
                    opacity=0.05
                ]
                    (-1.45,-1.45) rectangle (1.45,1.45);
                \draw[
                    gray!45,
                    line width=0.25pt,
                    step=2.9mm
                ]
                    (-1.45,-1.45) grid (1.45,1.45);
                \draw[
                    gray!70,
                    line width=0.5pt
                ]
                    (-1.45,-1.45) rectangle (1.45,1.45);
                \draw[
                    gray!70,
                    line width=0.5pt
                ]
                    (-1.45,0) -- (1.45,0)
                    (0,-1.45) -- (0,1.45);
                \ifdim\gridx pt=-3pt
                    \node[
                        anchor=south west,
                        font=\small
                    ]
                        at (-1.45,1.36) {$Q^{\mathsf T}z$};
                \fi
                \begin{axis}[
                    at={(0,0)},
                    anchor=center,
                    width=2.9cm,
                    height=2.9cm,
                    scale only axis,
                    xmin=-1.45,
                    xmax=1.45,
                    ymin=-1.45,
                    ymax=1.45,
                    axis equal image,
                    hide axis,
                    clip=false
                ]
                    \addplot[
                        SiennaBrand,
                        opacity=0.5,
                        line width=0.64pt,
                        smooth cycle
                    ]
                    table[
                        x=x_component,
                        y=y_component
                    ]{\fourierdata};
                \end{axis}
                \foreach \harmonic in {1,...,3}{
                    \ifnum\harmonic=1
                        \def\previousx{0}
                        \def\previousy{0}
                    \else
                        \pgfmathtruncatemacro{\previousharmonic}{\harmonic-1}
                        \edef\previousxcolumn{n\previousharmonic_cumulative_x}
                        \edef\previousycolumn{n\previousharmonic_cumulative_y}
                        \pgfplotstablegetelem{\datarow}{\previousxcolumn}\of\fourierdata
                        \edef\previousx{\pgfplotsretval}
                        \pgfplotstablegetelem{\datarow}{\previousycolumn}\of\fourierdata
                        \edef\previousy{\pgfplotsretval}
                    \fi
                    \edef\currentxcolumn{n\harmonic_cumulative_x}
                    \edef\currentycolumn{n\harmonic_cumulative_y}
                    \edef\magnitudecolumn{n\harmonic_magnitude}
                    \pgfplotstablegetelem{\datarow}{\currentxcolumn}\of\fourierdata
                    \edef\currentx{\pgfplotsretval}
                    \pgfplotstablegetelem{\datarow}{\currentycolumn}\of\fourierdata
                    \edef\currenty{\pgfplotsretval}
                    \pgfplotstablegetelem{\datarow}{\magnitudecolumn}\of\fourierdata
                    \edef\magnitude{\pgfplotsretval}
                    \draw[
                        CasalBrand!55,
                        dashed,
                        line width=0.4pt
                    ]
                        (\previousx,\previousy)
                        circle[radius=\magnitude cm];
                    \draw[
                        CasalBrand,
                        opacity=0.75,
                        line width=0.9pt
                    ]
                        (\previousx,\previousy)
                        -- (\currentx,\currenty);
                    \fill[
                        CasalBrand,
                        opacity=0.75
                    ]
                        (\currentx,\currenty) circle[radius=0.7pt];
                }
                \pgfplotstablegetelem{\datarow}{n3_cumulative_x}\of\fourierdata
                \edef\finalx{\pgfplotsretval}
                \pgfplotstablegetelem{\datarow}{n3_cumulative_y}\of\fourierdata
                \edef\finaly{\pgfplotsretval}
                \pgfplotstablegetelem{\pastrow}{n3_cumulative_x}\of\fourierdata
                \edef\pastx{\pgfplotsretval}
                \pgfplotstablegetelem{\pastrow}{n3_cumulative_y}\of\fourierdata
                \edef\pasty{\pgfplotsretval}
                \draw[
                    SiennaBrand,
                    line width=1.2pt,
                    -{Triangle[length=1mm,width=1mm]}
                ]
                    (0,0) -- (\finalx,\finaly);
            \end{scope}
        }
        \begin{scope}[rotate=45]
            \fill[black]
                (-3cm,-1.8mm)
                --
                (-3.11cm,1.2mm)
                --
                (-2.89cm,1.2mm)
                -- cycle;
        \end{scope}
        \foreach \startangle in {120,210,300,30}{
            \foreach \segment in {0,...,19}{
                \pgfmathsetmacro{\segmentstart}{
                    \startangle+1.5*\segment
                }
                \pgfmathsetmacro{\segmentend}{
                    \startangle+1.5*(\segment+1)
                }
                \pgfmathsetmacro{\segmentopacity}{
                    0.25+0.75*\segment/19
                }
                \ifnum\segment=19
                    \draw[
                        SiennaBrand,
                        opacity=\segmentopacity,
                        line width=0.9pt,
                        -{Triangle[length=1.5mm,width=1.2mm]}
                    ]
                        (\segmentstart:3.55cm)
                        arc[
                            start angle=\segmentstart,
                            end angle=\segmentend,
                            radius=3.55cm
                        ];
                \else
                    \draw[
                        SiennaBrand,
                        opacity=\segmentopacity,
                        line width=0.9pt
                    ]
                        (\segmentstart:3.55cm)
                        arc[
                            start angle=\segmentstart,
                            end angle=\segmentend,
                            radius=3.55cm
                        ];
                \fi
            }
        }
        \node[
            anchor=south east,
            font=\small,
            SiennaBrand
        ]
            at (135:3.55cm) {$R_{\pi/2}$};
        \node[
            anchor=north east,
            font=\small,
            SiennaBrand
        ]
            at (225:3.55cm) {$R_{\pi/2}$};
        \node[
            anchor=north west,
            font=\small,
            SiennaBrand
        ]
            at (315:3.55cm) {$R_{\pi/2}$};
        \node[
            anchor=south west,
            font=\small,
            SiennaBrand
        ]
            at (45:3.55cm) {$R_{\pi/2}$};
    \end{tikzpicture}
    \caption{\textbf{Fourier decomposition of a phase-equivariant latent
    trajectory.} Local Cartesian panels show the cumulative sum of the first
    three Fourier components at four cardinal phases. Dashed circles indicate
    component magnitudes, blue segments show the component-wise construction,
    orange arrows show the resultant latent vectors, and curved $R_{\pi/2}$
    arrows denote quarter-cycle group actions.}
    \label{fig:latent-space-rotation}
\end{figure*}

%% file: gapped_harmonics_resolution.tex

\pgfplotstableread[
    col sep=comma
]{data/figure_7_gapped_harmonics.csv}\gappedharmonicsdata

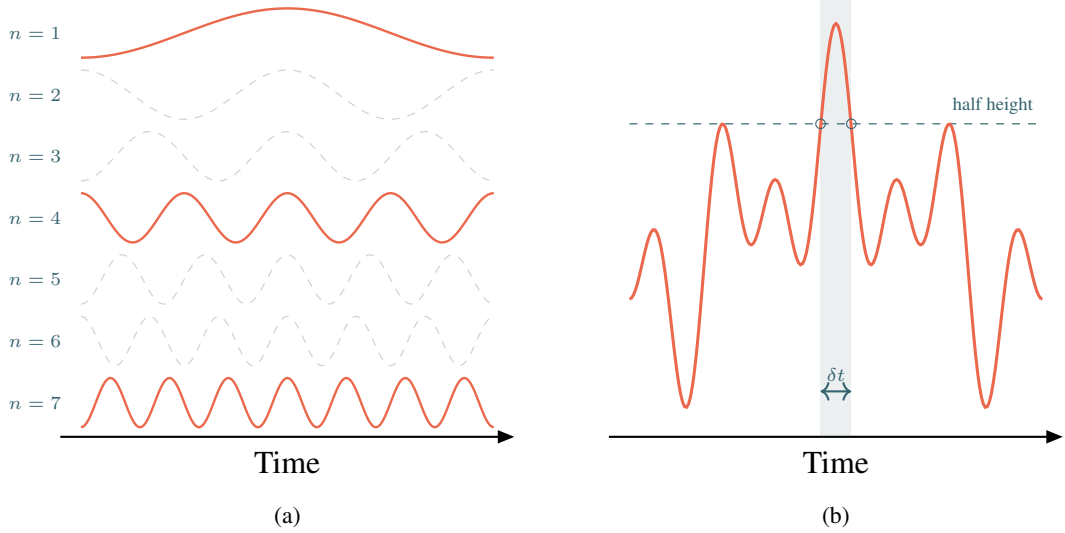
\begin{figure*}[!t]
    \centering
    \begin{tikzpicture}
        \begin{axis}[
            name=harmoniccomponents,
            width=0.43\textwidth,
            height=5.8cm,
            scale only axis,
            xmin=-0.05,
            xmax=1.05,
            ymin=-1,
            ymax=11.8,
            axis x line=bottom,
            axis y line=none,
            axis line style={
                -{Triangle[length=2.2mm,width=1.7mm]},
                line width=0.8pt,
                black
            },
            xtick=\empty,
            ytick=\empty,
            xlabel={Time},
            xlabel style={
                font=\large,
                text=black
            },
            clip=false
        ]
            \addplot[
                SiennaBrand,
                line width=0.9pt,
                no marks
            ]
                table[
                    x=phase_fraction,
                    y expr={\thisrow{n1}+10.8}
                ]{\gappedharmonicsdata};
            \addplot[
                black!18,
                dashed,
                line width=0.45pt,
                no marks
            ]
                table[
                    x=phase_fraction,
                    y expr={\thisrow{n2}+9.0}
                ]{\gappedharmonicsdata};
            \addplot[
                black!18,
                dashed,
                line width=0.45pt,
                no marks
            ]
                table[
                    x=phase_fraction,
                    y expr={\thisrow{n3}+7.2}
                ]{\gappedharmonicsdata};
            \addplot[
                SiennaBrand,
                line width=0.9pt,
                no marks
            ]
                table[
                    x=phase_fraction,
                    y expr={\thisrow{n4}+5.4}
                ]{\gappedharmonicsdata};
            \addplot[
                black!18,
                dashed,
                line width=0.45pt,
                no marks
            ]
                table[
                    x=phase_fraction,
                    y expr={\thisrow{n5}+3.6}
                ]{\gappedharmonicsdata};
            \addplot[
                black!18,
                dashed,
                line width=0.45pt,
                no marks
            ]
                table[
                    x=phase_fraction,
                    y expr={\thisrow{n6}+1.8}
                ]{\gappedharmonicsdata};
            \addplot[
                SiennaBrand,
                line width=0.9pt,
                no marks
            ]
                table[
                    x=phase_fraction,
                    y=n7
                ]{\gappedharmonicsdata};
            \node[anchor=east, xshift=-4pt, font=\scriptsize, text=CasalBrand]
                at (axis cs:0,10.8) {$n=1$};
            \node[anchor=east, xshift=-4pt, font=\scriptsize, text=CasalBrand]
                at (axis cs:0,9.0) {$n=2$};
            \node[anchor=east, xshift=-4pt, font=\scriptsize, text=CasalBrand]
                at (axis cs:0,7.2) {$n=3$};
            \node[anchor=east, xshift=-4pt, font=\scriptsize, text=CasalBrand]
                at (axis cs:0,5.4) {$n=4$};
            \node[anchor=east, xshift=-4pt, font=\scriptsize, text=CasalBrand]
                at (axis cs:0,3.6) {$n=5$};
            \node[anchor=east, xshift=-4pt, font=\scriptsize, text=CasalBrand]
                at (axis cs:0,1.8) {$n=6$};
            \node[anchor=east, xshift=-4pt, font=\scriptsize, text=CasalBrand]
                at (axis cs:0,0) {$n=7$};
        \end{axis}

        \begin{axis}[
            name=harmonicsum,
            at={(harmoniccomponents.south east)},
            anchor=south west,
            xshift=1.25cm,
            width=0.43\textwidth,
            height=5.8cm,
            scale only axis,
            xmin=-440,
            xmax=440,
            ymin=-1.0,
            ymax=1.12,
            axis x line=bottom,
            axis y line=none,
            axis line style={
                -{Triangle[length=2.2mm,width=1.7mm]},
                line width=0.8pt,
                black
            },
            xtick=\empty,
            xlabel={Time},
            xlabel style={
                font=\large,
                text=black
            },
            clip=false
        ]
            \path[
                fill=CasalBrand,
                opacity=0.10,
                draw=none
            ]
                (axis cs:-29.85,-1.0)
                rectangle
                (axis cs:29.85,1.12);
            \addplot[
                CasalBrand,
                dashed,
                line width=0.45pt
            ]
                coordinates {
                    (-400,0.515)
                    (400,0.515)
                };
            \node[
                anchor=south east,
                font=\scriptsize,
                text=CasalBrand
            ]
                at (axis cs:400,0.515) {half height};
            \addplot[
                SiennaBrand,
                line width=1.15pt,
                no marks
            ]
                table[
                    x=time_ms,
                    y=sum_1_4_7
                ]{\gappedharmonicsdata};
            \addplot[
                only marks,
                mark=o,
                mark size=1.8pt,
                CasalBrand,
                fill=white
            ]
                coordinates {
                    (-29.85,0.515)
                    (29.85,0.515)
                };
            \draw[
                <->,
                CasalBrand,
                line width=0.8pt
            ]
                (axis cs:-29.85,-0.78)
                --
                node[above, font=\scriptsize] {$\delta t$}
                (axis cs:29.85,-0.78);
        \end{axis}
        \node[
            anchor=north,
            font=\small
        ]
            at ([yshift=-0.8cm]harmoniccomponents.south) {(a)};
        \node[
            anchor=north,
            font=\small
        ]
            at ([yshift=-0.8cm]harmonicsum.south) {(b)};
    \end{tikzpicture}
    \caption{\textbf{Localisation from a gapped harmonic band.}
    \textbf{(a)} Harmonics $n=1,\ldots,7$, with the retained blocks
    $\{1,4,7\}$ shown in orange. \textbf{(b)} Their normalised sum over a
    cardiac cycle. The shaded full width at half maximum, $\delta t$, denotes
    the temporal scale that this harmonic band can resolve, while the gaps
    leave pronounced side lobes.}
    \label{fig:gapped-harmonic-resolution}
\end{figure*}

%% file: method.tex

\section{Method}
\label{sec:method}

\subsection{Dataset}

The PTB-XL dataset \citep{wagner2020ptb,strodthoff2021ptbxlbench} was selected
because its acquisition and annotation structure match both the mechanism
under study and its downstream evaluation. It provides fixed-length,
10-second, 12-lead clinical ECGs at 500\,Hz, giving sufficient temporal
resolution to construct a cardiac-phase coordinate from R-peak timing.
Its patient-aware stratified folds and hierarchical diagnostic labels provide a reproducible route from self-supervised pre-training to diagnostic evaluation. PTB-XL therefore provides an established ECG dataset whose structure directly supports the phase-indexed latent space learning problem.

Version 1.0.3 of the corpus comprises 21,799 recordings from 18,869 patients
\citep{PhysioNet-ptb-xl-1.0.3}: clinical resting ECGs, each 10\,s long and
sampled at 500\,Hz across the 12 standard leads. Each recording carries one or
more of 71 SCP-ECG\footnote{`Standard Communications Protocol for
Computer-Assisted Electrocardiography'.} statements, assigned by cardiologists
and spanning diagnostic, form and rhythm categories.

Classification of the PTB-XL diagnostic subclasses provides the primary
evaluation of the learned latent representations. The ``23-subclass'' metric
averages performance across the 17 classes with at least 20 positive fold-9
examples. The five-group macro-average follows PTB-XL's published diagnostic
hierarchy, which maps these labels into normal ECG, myocardial infarction,
ST/T change, conduction disturbance, and hypertrophy superclasses.


The corpus is used in two stages, separated so that nothing reported on the
sealed fold was chosen with reference to it. In the development stage, models
are fitted on folds 1--8 and scored on fold 9. That stage fixes the design
choices by grid search: the number of encoder layers, the number of predictor
layers, and the loss-weight prefactors of \cref{eq:winder-total-loss}. It also
fixes the isotypic capacity allocation, meaning the split of the
256-dimensional latent budget between the invariant capacity $K_0$ and the
harmonic multiplicities $\{c_n\}$ of \cref{eq:latent-capacity}. The highest
retained harmonic $n_{\max}$ is not tuned in this stage; it is set in advance
from the physiological resolution requirement of
\cref{sec:harmonic-spectrum}. Fold 9 is the only partition consulted in making
any of these choices, and it is likewise where the eligible subclass list is
defined.

In the confirmatory stage, those settings are frozen and the model is
pre-trained on the patient-disjoint folds 1--9 (19,601 recordings; 16,965
patients), over which the per-lead normalisation statistics are also fitted. A
linear probe on the frozen representation is fitted over the same folds and
evaluated on fold 10 (2,198 recordings; 1,904 patients), which remained sealed
until unblinding; results reported on it are this paper's headline evaluation.
The two partitions agree to within a percentage point on age, sex and every
class prevalence (\cref{tab:cohort-composition}), which is what PTB-XL's
stratification should deliver and what was verified rather than assumed.

Recordings are read at the native 500\,Hz and decimated locally to 100\,Hz by
polyphase FIR resampling, giving 1,000 samples per lead. The whole recording
is retained as a single training example. Each lead is then $z$-scored against per-lead corpus statistics fitted once over folds 1--9.

The encoder receives the resampled 100\,Hz signal, whereas R-peaks are detected once on the native 500\,Hz recordings with a Pan--Tompkins-style detector \citep{pan1985real}.
Cardiac phase is then estimated by linear interpolation between consecutive peaks, as in \cref{eq:phase-map}.
Each 80\,ms token is assigned the phase at the centre of its patch.

Samples before the first detected peak and at or after the final detected peak have undefined phase, and are excluded from phase-dependent analyses. \cref{app:data} states the full preprocessing chain and the fitted statistics.

\subsection{Model architecture}
\input{method-architecture-figure}

We implement \Winder{} as a causal joint-embedding predictive architecture for
10-second, 12-lead ECGs. Each recording is represented by 125 non-overlapping
tokens, each corresponding to 80\,ms of signal sampled at 100\,Hz. The distinguishing component
is \emph{not} the JEPA framework itself, but the fixed phase-indexed transport
operator that constrains the projected representation to transform according
to cardiac phase (cf. \cref{eq:canonical-block-decomposition}).
\Cref{fig:winder-architecture} shows the complete JEPA architecture,
distinguishing the deployed path from components used only during training.

The encoder partitions each input into 125 non-overlapping eight-sample
patches. Each $12\times8$ patch is flattened to 96 values and passed through
the same per-patch MLP, with dimensions $96\rightarrow512\rightarrow256$ and
a GELU nonlinearity. The resulting 256-dimensional tokens pass through two
residual convolutional blocks with kernel size 3 and dilations 1 and 2 respectively.
Left-only padding preserves causality. Across the four convolutions, each
output depends on the current token and the preceding 12 tokens, giving a
0.96\,s look-back.

Within each context block, the residual branch applies channel-wise LayerNorm,
GELU, and causal convolution twice beneath a single identity connection. No
normalisation or activation follows the residual addition, and the projector
output is likewise left unnormalised. This allows SIGReg to control the scale
and moments of the projected distribution, which a terminal LayerNorm would
otherwise fix by construction.

A shared projector with dimensions $256\rightarrow512\rightarrow256$ and a
GELU nonlinearity independently maps each of the 125 contextual encoder tokens
to $z_t\in\mathbb{R}^{256}$, without mixing information across time. These
tokens support the self-supervised training losses. Together, the encoder and projector form the deployed representation path and contain 1,232,384 parameters.
For downstream evaluation, tokens with defined cardiac phase are mean-pooled
without weighting to form a record-level representation.

The selected transport operator instantiates the block decomposition of
\cref{eq:canonical-block-decomposition} in the 256-dimensional projected
space. The invariant capacity is $K_0=4$, and the remaining 252 dimensions
carry harmonics $n=1,\ldots,10$ with multiplicities
$(c_1,\ldots,c_{10})=(24,24,20,16,12,10,8,6,4,2)$, each unit of $c_n$
contributing one two-dimensional block, so that $2\sum_n c_n = 252$. Each
coordinate pair in harmonic $n$ undergoes a closed-form rotation by $n\Delta$.
The operator is fixed and has zero learnable parameters.

During self-supervised training, the projected tokens also enter a causal
Transformer predictor comprising four pre-LayerNorm blocks, four attention
heads, width 256, and feed-forward width 1,024. It uses learned
relative-position biases. The predictor adds 3,227,084
parameters, increasing the optimised total from the deployed 1,232,384
parameters to 4,459,468. It is discarded after pre-training and contributes
neither to deployment nor to the reported diagnostic probes. The masking and
prediction-loss mechanics are described in \cref{sec:training}.

Latent-geometry analyses use the pre-projection encoder output to test whether
phase structure is learned by the encoder itself. Training losses and
diagnostic probes instead use the projected tokens $z_t$, because these are the
representations directly constrained by the objectives and used for downstream
evaluation.
For diagnostic evaluation, the mean-pooled representation is passed to a linear
probe fitted separately for each task. This probe is an evaluation readout,
not part of the pretrained model, and its weights are excluded from the
reported parameter counts.

\subsection{Training}
\label{sec:training}

Self-supervised training minimises the three-term objective of \cref{eq:winder-total-loss}, with all three terms attached to the projector output $z$. The live weights are $\lambda_{\mathrm{SIG}}=0.15$ and $\lambda_{\mathrm{trans}}=1$. The nominal loss weights were selected through a coarse hyperparameter sweep. In the control condition, the transport loss is disabled by setting $\lambda_{\mathrm{trans}}=0$.

For each record, the prediction loss applies only to the token after a sampled causal cutoff (\cref{sec:latent-prediction}). The cutoff is drawn independently per record, uniformly over admissible token indices, and redrawn every batch. Inputs after the cutoff are replaced by a learned mask token, so the output at $c+1$ depends only on the visible latent prefix and is compared with $z_{c+1}$. The evaluation protocol of the SIGReg loss is given in \cref{app:training-sigreg}. The transport loss, $\mathcal{L}_{\mathrm{trans}}$, is averaged over all valid token pairs within each record, and then across records. Using all valid token pairs avoids the phase-dependent sampling bias observed when only a subset of pairs was selected.

Training uses AdamW with a short linear learning-rate warm-up followed by
cosine decay. The optimiser and schedule are fixed across all runs. Their
complete specification, including batch size and training duration, is given
in \cref{app:training-optim}.

During training, each of six signal augmentations is applied independently
with probability 50 \%. Gaussian noise and a 25 Hz sinusoid simulate broadband
and narrow-band measurement noise, respectively. Low-frequency drift models
baseline wander, while slow amplitude modulation represents gradual changes
in signal strength. Lead dropout simulates missing channels, and per-lead gain
scaling varies the relative amplitudes between leads.
Full augmentation details are reported in \cref{app:training-augment}.

Four pre-training runs were conducted: a transport-enabled arm and a control
arm, each using two random seeds.
The complete run design, checkpoint schedule, and hardware specification are given in
\cref{app:training-runs}.

%% file: method-architecture-figure.tex
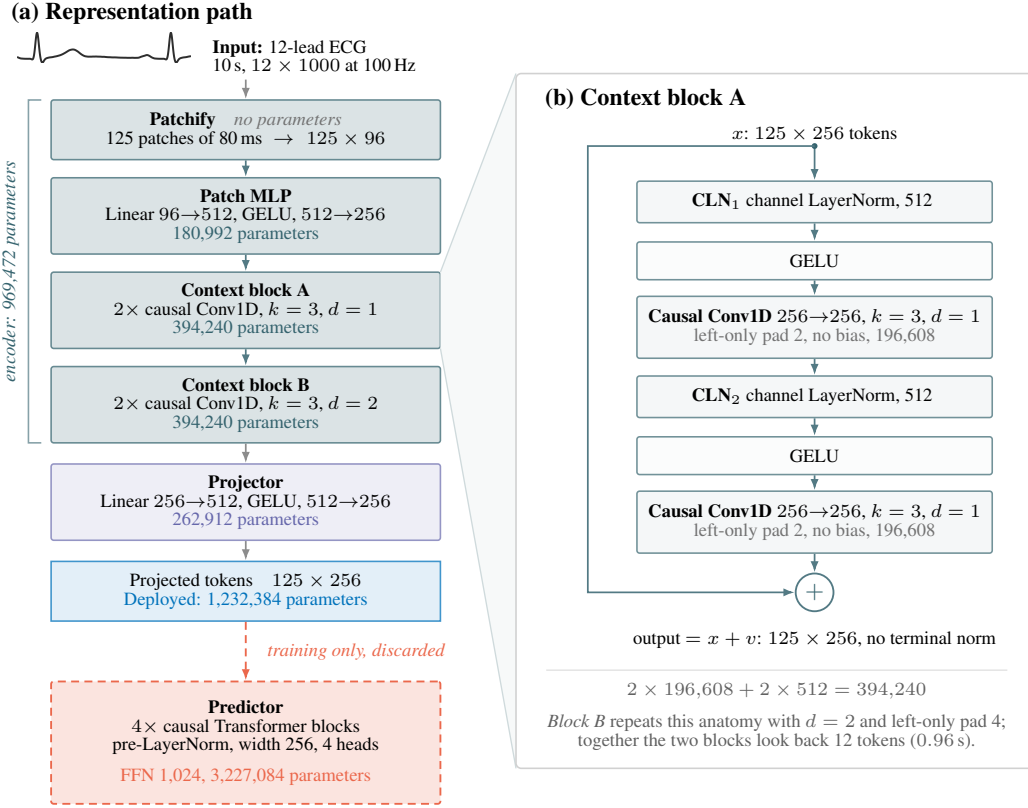
\begin{figure*}[!t]
    \centering
    \begin{tikzpicture}[
        x=1cm,
        y=1cm,
        deployedflow/.style={
            -{Latex[length=1.5mm,width=1.1mm]},
            draw=CasalBrand!85,
            line width=0.65pt
        },
        fixedflow/.style={
            -{Latex[length=1.5mm,width=1.1mm]},
            draw=black!45,
            line width=0.65pt
        },
        trainingflow/.style={
            -{Latex[length=1.5mm,width=1.1mm]},
            draw=SiennaBrand,
            densely dashed,
            line width=0.7pt
        },
        stageblock/.style={
            draw=CasalBrand!75,
            fill=CasalBrand!18,
            rounded corners=1pt,
            line width=0.6pt,
            minimum width=5.15cm,
            inner xsep=1.4mm,
            inner ysep=0pt,
            align=center,
            font=\scriptsize
        },
        neutralstage/.style={
            stageblock,
            draw=ProjectorAccent!70,
            fill=ProjectorAccent!11
        },
        outputstage/.style={
            stageblock,
            draw=DeployedAccent!75,
            fill=DeployedAccent!11,
            sharp corners
        },
        predictorstage/.style={
            stageblock,
            draw=SiennaBrand,
            fill=SiennaBrand!18,
            densely dashed,
            line width=0.7pt
        },
        anatomyblock/.style={
            draw=CasalBrand!70,
            fill=CasalBrand!5,
            rounded corners=1pt,
            line width=0.6pt,
            minimum width=4.70cm,
            inner xsep=1.2mm,
            inner ysep=0pt,
            align=center,
            font=\scriptsize
        },
        paneltitle/.style={
            anchor=west,
            font=\small
        },
        sidelabel/.style={
            rotate=90,
            anchor=south,
            text=CasalBrand,
            font=\scriptsize\itshape
        },
        panelnote/.style={
            align=center,
            text=black!62,
            font=\scriptsize
        }
    ]
        \path[use as bounding box] (0,-0.55) rectangle (13.80,10.15);

        \node[paneltitle] at (0.10,10.00) {\textbf{(a) Representation path}};

        \begin{axis}[
            at={(0.30cm,9.345cm)},
            anchor=south west,
            width=2.30cm,
            height=0.395cm,
            scale only axis,
            axis lines=none,
            xmin=-0.126,
            xmax=0.9653,
            ymin=-0.1618115238,
            ymax=0.5700201847,
            clip=true,
        ]
            \addplot[
                black!82,
                line width=0.75pt,
            ]
            table[
                col sep=comma,
                x=time_s,
                y=amplitude,
            ]{data/phase_wrap_time.csv};
        \end{axis}
        \node[anchor=west, align=left, font=\scriptsize] at (2.76,9.42) {
            \textbf{Input:} 12-lead ECG\\[-1pt]
            10\,s, $12\times1000$ at 100\,Hz
        };

        \node[stageblock, minimum height=0.79cm] (patchify) at (3.325,8.455) {
            \textbf{Patchify}\quad
            \textcolor{black!55}{\itshape no parameters}\\[-1pt]
            125 patches of 80\,ms
            $\;\rightarrow\;125\times96$
        };

        \node[stageblock, minimum height=1.01cm] (patchmlp) at (3.325,7.315) {
            \textbf{Patch MLP}\\[-1pt]
            Linear $96{\to}512$, GELU, $512{\to}256$\\[-1pt]
            \textcolor{CasalBrand}{180,992 parameters}
        };

        \node[stageblock, minimum height=1.01cm] (contexta) at (3.325,6.065) {
            \textbf{Context block A}\\[-1pt]
            $2\times$ causal Conv1D, $k=3$, $d=1$\\[-1pt]
            \textcolor{CasalBrand}{394,240 parameters}
        };

        \node[stageblock, minimum height=1.01cm] (contextb) at (3.325,4.815) {
            \textbf{Context block B}\\[-1pt]
            $2\times$ causal Conv1D, $k=3$, $d=2$\\[-1pt]
            \textcolor{CasalBrand}{394,240 parameters}
        };

        \node[neutralstage, minimum height=1.01cm] (projector) at (3.325,3.525) {
            \textbf{Projector}\\[-1pt]
            Linear $256{\to}512$, GELU, $512{\to}256$\\[-1pt]
            \textcolor{ProjectorAccent}{262,912 parameters}
        };

        \node[outputstage, minimum height=0.79cm]
            (projected) at (3.325,2.345) {
            Projected tokens\quad $125\times256$\\[-1pt]
            \textcolor{DeployedAccent}{Deployed: 1,232,384 parameters}
        };

        \node[predictorstage, minimum height=1.62cm] (predictor)
            at (3.325,0.34) {
            \textbf{Predictor}\\[-1pt]
            $4\times$ causal Transformer blocks\\[-1pt]
            pre-LayerNorm, width 256, 4 heads\\[3.5pt]
            \textcolor{SiennaBrand}{FFN 1,024, 3,227,084 parameters}
        };

        \draw[fixedflow] (3.325,9.12) -- (patchify.north);
        \draw[deployedflow] (patchify.south) -- (patchmlp.north);
        \draw[deployedflow] (patchmlp.south) -- (contexta.north);
        \draw[deployedflow] (contexta.south) -- (contextb.north);
        \draw[fixedflow] (contextb.south) -- (projector.north);
        \draw[fixedflow] (projector.south) -- (projected.north);
        \draw[trainingflow] (projected.south) --
            node[
                anchor=west,
                xshift=1.6mm,
                text=SiennaBrand,
                font=\scriptsize\itshape
            ] {training only, discarded}
            (predictor.north);

        \draw[CasalBrand!60, line width=0.6pt]
            (0.62,8.85) -- (0.45,8.85) -- (0.45,4.31) -- (0.62,4.31);
        \node[sidelabel] at (0.45,6.58) {encoder: 969,472 parameters};

        \path[fill=CasalBrand!4]
            (5.90,6.57) -- (6.90,9.20) -- (6.90,0.00) -- (5.90,5.56) -- cycle;
        \draw[CasalBrand!24, line width=0.5pt] (5.90,6.57) -- (6.90,9.20);
        \draw[CasalBrand!24, line width=0.5pt] (5.90,5.56) -- (6.90,0.00);

        \draw[black!18, rounded corners=2pt, line width=0.6pt]
            (6.90,0.00) rectangle (13.80,9.20);
        \node[paneltitle] at (7.16,8.87) {\textbf{(b) Context block A}};
        \node[font=\scriptsize] at (10.85,8.40)
            {$x$: $125\times256$ tokens};

        \node[anatomyblock, minimum height=0.55cm] (clnone) at (10.85,7.50) {
            \textbf{CLN$_1$} channel LayerNorm, 512
        };
        \node[anatomyblock, fill=CasalBrand!2, minimum height=0.50cm]
            (geluone) at (10.85,6.72) {GELU};
        \node[anatomyblock, minimum height=0.82cm] (convone) at (10.85,5.84) {
            \textbf{Causal Conv1D} $256{\to}256$, $k=3$, $d=1$\\[-1pt]
            \textcolor{black!55}{left-only pad 2, no bias, 196,608}
        };
        \node[anatomyblock, minimum height=0.55cm] (clntwo) at (10.85,4.92) {
            \textbf{CLN$_2$} channel LayerNorm, 512
        };
        \node[anatomyblock, fill=CasalBrand!2, minimum height=0.50cm]
            (gelutwo) at (10.85,4.14) {GELU};
        \node[anatomyblock, minimum height=0.82cm] (convtwo) at (10.85,3.26) {
            \textbf{Causal Conv1D} $256{\to}256$, $k=3$, $d=1$\\[-1pt]
            \textcolor{black!55}{left-only pad 2, no bias, 196,608}
        };
        \node[
            circle,
            draw=CasalBrand!85,
            fill=white,
            line width=0.7pt,
            minimum size=5mm,
            inner sep=0pt,
            text=CasalBrand,
            font=\small
        ] (residualadd) at (10.85,2.33) {$+$};

        \draw[deployedflow] (10.85,8.24) -- (clnone.north);
        \draw[deployedflow] (clnone.south) -- (geluone.north);
        \draw[deployedflow] (geluone.south) -- (convone.north);
        \draw[deployedflow] (convone.south) -- (clntwo.north);
        \draw[deployedflow] (clntwo.south) -- (gelutwo.north);
        \draw[deployedflow] (gelutwo.south) -- (convtwo.north);
        \draw[deployedflow] (convtwo.south) -- (residualadd.north);

        \fill[CasalBrand] (10.85,8.24) circle (1.1pt);
        \draw[deployedflow]
            (10.85,8.24) -- (7.85,8.24) -- (7.85,2.33) -- (residualadd.west);
        \node[font=\scriptsize] at (10.85,1.70)
            {output $=x+v$: $125\times256$, no terminal norm};
        \draw[black!12, line width=0.5pt] (7.30,1.30) -- (13.40,1.30);
        \node[panelnote] at (10.35,1.02)
            {$2\times196{,}608 + 2\times512 = 394{,}240$};
        \node[panelnote] at (10.35,0.48) {
            \textit{Block B} repeats this anatomy with $d=2$ and left-only
            pad 4;\\[-1pt]
            together the two blocks look back 12 tokens ($0.96$\,s).
        };
    \end{tikzpicture}
    \caption{\textbf{\Winder{} architecture.}
    \textbf{(a)} The encoder maps 80 ms of 12-lead ECG to a 256-dimensional latent space. Fixed patchification, a patch MLP, and two causal context blocks form the 969,472-parameter \textcolor{CasalBrand}{encoder}. The \textcolor{ProjectorAccent}{projector} produces 256-dimensional latent tokens from the encoder's output. Encoder and projector together form the 1,232,384-parameter \textcolor{DeployedAccent}{deployed path}.
    During self-supervised training the projected tokens also enter the 3,227,084-parameter pre-LayerNorm causal Transformer \textcolor{SiennaBrand}{predictor}, which is \emph{discarded at deployment}.
    \textbf{(b)} Context block A repeats layer normalisation, GELU, and causal convolution twice under one identity skip, with no normalisation after the residual addition.
    Left-only padding in the convolutions ensures causality.
    }
    \label{fig:winder-architecture}
\end{figure*}

%% file: 04_results.tex

\section{Results}
\label{sec:results}

Published ECG representation-learning results vary in their pre-training data,
parameter scale, input windowing, and downstream readout. A matched ablation
therefore isolates the contribution of phase transport. The two arms of the ablation
share their data, architecture, random seeds, training schedule, and evaluation protocol. They differ only in the transport objective: \Winder{} sets
$\lambda_{\mathrm{trans}}=1.0$, whereas the control disables phase transport
by setting $\lambda_{\mathrm{trans}}=0.0$. 
This comparison evaluates the effect of phase transport on downstream
performance, use of the declared phase action, and any associated robustness
gains or failure modes.
Because phase transport acts only during training, both arms have identical
inference paths.

The main rows of \cref{tab:bottomline} report seed~0 at step~5{,}000 on sealed
fold~10. The training step was selected after the results were known:
step~5{,}000 has the largest seed-averaged superclass gap and the smallest
seed-averaged lead-dropout separation. 
The pre-registered evaluation grid comprised all 12 combinations of two seeds,
three training steps, and two arms. 
Across that grid, the seed-averaged superclass gap ranges from $+0.0774$ to $+0.0910$. 
None of the six matched cross-arm confidence-interval pairs overlaps.
The two seeds agree: the largest between-seed spread in any
\Winder{} superclass cell is $0.0054$, well inside the confidence intervals.
These results provide within-study evidence from one sealed PTB-XL fold. They
do not establish generalisation to external datasets or a directly comparable
leaderboard ranking.

\begin{table*}[!t]
  \centering
  \small
  \setlength{\tabcolsep}{4pt}
  \renewcommand{\arraystretch}{1.08}
  \newcommand{\resultsection}[1]{%
    \tikz[baseline=(section.base)]{%
      \node[
        anchor=base west,
        fill=CasalBrand!18,
        rounded corners=2pt,
        inner xsep=4pt,
        inner ysep=2pt,
        text width=\dimexpr0.88\textwidth+4\tabcolsep-8pt\relax
      ] (section) {\textit{#1}};%
    }%
  }
  \begin{tabular}{@{}p{0.45\textwidth}p{0.215\textwidth}p{0.215\textwidth}@{}}
    \toprule
    Metric &
    \Winder{} ($\lambda_{\mathrm{trans}}{=}1.0$) &
    Control ($\lambda_{\mathrm{trans}}{=}0.0$) \\
    \midrule

    \multicolumn{3}{@{}l@{}}{\resultsection{Diagnostic accuracy}} \\
    Superclass macro-AUROC, 5 classes $\uparrow$ &
    $\bm{0.8702^{+0.0092}_{-0.0095}}$ &
    $0.7826^{+0.0115}_{-0.0115}$ \\
    Subclass macro-AUROC, 17 classes $\uparrow$ &
    $\bm{0.8421^{+0.0105}_{-0.0120}}$ &
    $0.7568^{+0.0144}_{-0.0161}$ \\
    Subclass macro-AUROC, 23 classes $\uparrow$ &
    $\bm{0.8236^{+0.0220}_{-0.0197}}$ &
    $0.7415^{+0.0284}_{-0.0289}$ \\
    \addlinespace

    \multicolumn{3}{@{}l@{}}{\resultsection{Mechanism validity}} \\
    Transport gain fraction $\uparrow$ &
    \textbf{0.8786} &
    $-0.0744$ \\
    G1 shuffled-$\phi$ gate &
    \textbf{Pass} &
    Fail \\
    Paired true $-$ shuffled gain $\uparrow$ &
    $\bm{0.8154^{+0.0159}_{-0.0177}}$ &
    $0.0325^{+0.0023}_{-0.0024}$ \\
    \addlinespace

    \multicolumn{3}{@{}l@{}}{\resultsection{Robustness to degradation}} \\
    Mean signed $\Delta$ macro-AUROC, single-lead dropout $\uparrow$ &
    $\bm{-0.0034}$ &
    $-0.0107$ \\
    Worst signed $\Delta$ macro-AUROC across single-lead dropouts $\uparrow$ &
    $\bm{-0.0089}$ (aVR) &
    $-0.0218$ (III) \\
    \addlinespace

    \multicolumn{3}{@{}l@{}}{\resultsection{Latent geometry}} \\
    Fundamental loop amplitude at $\sigma=0$ $\uparrow$ &
    \textbf{1.8941} &
    0.3817 \\
    Fundamental-loop coherence (descriptive), mean over the
    7 non-reference bins $\uparrow$ &
    \textbf{0.9927} &
    0.3159 \\
    \addlinespace

    \multicolumn{3}{@{}l@{}}{\resultsection{Anomaly detection
    (strongest of six families; one fixed detector, maximum severity)}} \\
    Ectopic-beat AUROC, severity 2.0 $\uparrow$ &
    \textbf{0.8662} &
    0.6198 \\
    Excess over own zero-severity reference $\uparrow$ &
    $\bm{+0.2758}$ &
    $+0.1393$ \\
    Localisation hit rate $\uparrow$ &
    \textbf{0.779} &
    0.025 \\
    Median latency among alarmed records $\downarrow$ &
    \textbf{80\,ms} &
    720\,ms \\



    \bottomrule
  \end{tabular}
  \caption{\textbf{\Winder{} ($\lambda_{\mathrm{trans}}=1.0$) vs. control ($\lambda_{\mathrm{trans}}=0.0$).} Phase transport improves diagnostic accuracy, latent
  organisation, and lead-dropout robustness at identical inference cost.
  95\,\% CIs are patient-clustered bootstraps.
  Intervals are 1{,}000-replicate percentile bounds; remaining values are
  point estimates. Arrows give the preferred direction; bold marks the
  better arm where the comparison is non-degenerate. Provenance, protocol,
  and the caveats attaching to each block are given in the text.}
  \label{tab:bottomline}
\end{table*}

\subsection{Diagnostic accuracy.}
Each frozen representation is evaluated with the same linear-probe protocol,
with uncertainty estimated by resampling patients rather than records over
1{,}000 bootstrap replicates. Sealed fold~10 holds 2{,}198 records, of which
2{,}158 carry a diagnostic superclass. The probe scores 2{,}140 of these; 18 are excluded because they lack a valid phase assignment after quality control. At the
selected cell, \Winder{} reaches superclass macro-AUROC
$0.8702^{+0.0092}_{-0.0095}$, compared with $0.7826^{+0.0115}_{-0.0115}$ for
the control. The corresponding 17-class scores are
$0.8421^{+0.0105}_{-0.0120}$ and $0.7568^{+0.0144}_{-0.0161}$; the 23-class
scores are $0.8236^{+0.0220}_{-0.0197}$ and $0.7415^{+0.0284}_{-0.0289}$.
Phase transport therefore makes diagnostic structure more
linearly accessible on the sealed fold. This performance claim is specific to
a frozen linear readout. The post-hoc heart-rate stratification reported below
weakens, but does not exclude, an explanation of this gap based on heart rate.
No published result shares this protocol exactly, so \Cref{app:ptbxl} places
\Winder{} within the range reported by self-supervised models on PTB-XL, at a
smaller parameter and pre-training footprint, rather than ranking it against
them.

\subsection{Mechanism validity.}

For tokens at $t$ and $t'$, define the transported and untransported cosine
similarities as
\begin{equation}
\begin{aligned}
c_R &\coloneqq
\left\langle R_{\Delta_{tt'}}\hat{z}_t,\hat{z}_{t'}\right\rangle, \\
c_I &\coloneqq
\left\langle \hat{z}_t,\hat{z}_{t'}\right\rangle.
\end{aligned}
\label{eq:transport-gain}
\end{equation}
The raw pairwise gain is $g_{tt'}=c_R-c_I$. Its mean and the reported
gain fraction are, respectively,
\begin{equation}
\bar{g}\coloneqq\mathbb{E}_{tt'}[c_R-c_I],
\qquad
g_{\mathrm f}\coloneqq
\frac{\mathbb{E}_{tt'}[c_R-c_I]}
{\mathbb{E}_{tt'}[1-c_I]}.
\label{eq:transport-gain-summary}
\end{equation}
Thus, the gain fraction normalises the mean gain by the available improvement
from the untransported similarity to unit similarity.
The pre-registered G1 gate compares this
quantity under the true cardiac clock with the result after shuffling the phase
labels. It requires the paired true-minus-shuffled interval to exclude zero and
the shuffled gain fraction to satisfy $|g_{\mathrm{f}}|\leq0.02$. \Winder{}
achieves mean gain $\bar g=0.8182$ and gain fraction $g_{\mathrm f}=0.8786$,
and passes G1; the control's gain fraction is $-0.0744$ and it fails.

The paired true-minus-shuffled gains are $0.8154^{+0.0159}_{-0.0177}$ for
\Winder{} and $0.0325^{+0.0023}_{-0.0024}$ for the control. The control
interval also excludes zero, so
the shuffled-magnitude condition carries the arm discrimination: its shuffled
gain fraction is $-0.1104$, outside the registered tolerance. The gain fraction
itself is a point estimate; the available interval brackets mean gain
($0.8182^{+0.0154}_{-0.0173}$ for \Winder{}) and cannot be attached to the fraction. This
test establishes that the constrained representation uses the declared cardiac
clock; it does not by itself explain the diagnostic gain.

\begin{figure*}[!t]
  \centering
  \makebox[\textwidth][c]{%
    \includegraphics[width=1.06\textwidth]{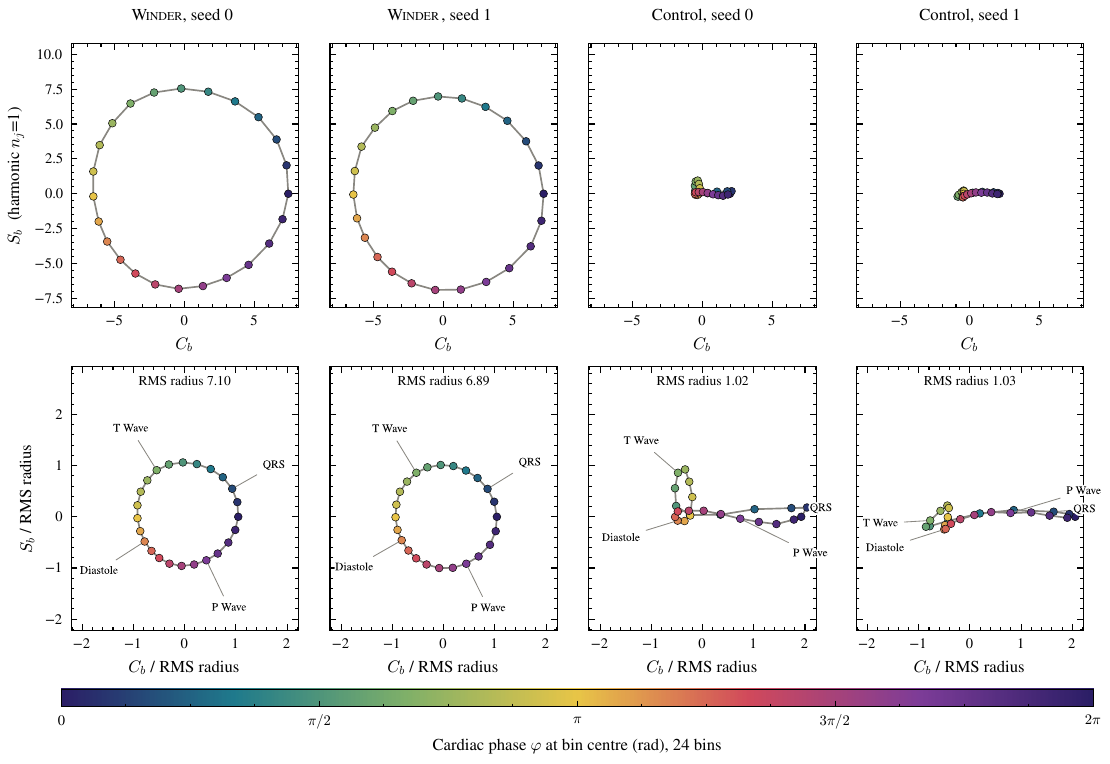}}
  \caption{\textbf{Latent cardiac-phase organisation with and without
  transport.} Visualisation of the cosine and sine overlaps defined in
  \cref{eq:coherence-overlaps}. Tokens are grouped into 24 cardiac-phase bins;
  each point plots $(C_b,S_b)$ for the mean $n=1$ representation of one bin
  relative to the reference bin. Each column shows one run: both \Winder{}
  seeds and both control seeds. Colour encodes $\phi$ cyclically, so the first
  and last bins are neighbours in colour as well as in phase. \emph{Top row:} shared axis limits, so loop size is comparable across panels. \emph{Bottom row:} each
  panel normalised by its own RMS radius, so loop shape is comparable. To visualise the same overlap construction at finer phase resolution, tokens from folds~1--9 are grouped into 24 cardiac-phase bins; the sealed-fold coherence estimate uses eight bins.}
  \label{fig:phase-ring}
\end{figure*}

\subsection{Single-lead robustness.}
Each lead is zeroed independently at evaluation while the probe remains fitted
on intact data. The resulting AUROC is compared with that arm's own intact
score. The mean signed change is $-0.0034$ for \Winder{} and $-0.0107$ for the
control, a $3.1\times$ ratio at step~5{,}000. The worst signed changes are
$-0.0089$ after removing aVR and $-0.0218$ after removing lead~III,
respectively. The seed-averaged mean-degradation ratio widens to $11.3\times$
by step~30{,}000. These robustness summaries are descriptive point estimates.


\subsection{Latent geometry.}
Two complementary metrics quantify phase organisation in latent space.
\emph{Fundamental-harmonic Amplitude} measures how strongly token representations
align after removing their cardiac phase, while \emph{Orbit-subspace Coherence}
measures whether phase-bin means remain within a common rotational subspace. 

For a token $t$ with cardiac phase $\phi_t$, the dephased latent is defined as
\begin{equation}
\widetilde z_t \coloneqq R_{-\phi_t}z_t.
\label{eq:dephased-latent}
\end{equation}
This counter-rotation brings phase-locked components to the common reference
phase $\phi=0$. To calculate the Fundamental-harmonic Amplitude, first average the dephased latents per record. Then, compute the mean Euclidean norm of the 24 two-dimensional $n=1$ components. The resulting amplitude is $1.8941$ for \Winder{} and $0.3817$ for the control, indicating a roughly fivefold stronger phase-locked fundamental component in \Winder{}.

\emph{Orbit-subspace coherence} is computed from the original, non-dephased
token representations. Tokens are grouped into eight cardiac-phase bins, and
the mean $n=1$ representation $v_b\in\mathbb{R}^{2c_1}$ is calculated for each
bin. Let
\begin{equation}
u\coloneqq\frac{v_0}{\lVert v_0\rVert_2}
\label{eq:coherence-reference}
\end{equation}
denote the normalised mean of the reference bin, and let
\begin{equation}
J\coloneqq
I_{c_1}\otimes
\begin{pmatrix}
0&-1\\
1&0
\end{pmatrix}
\label{eq:coherence-quarter-turn}
\end{equation}
apply a simultaneous quarter-turn to all $c_1$ two-dimensional blocks. The
cosine and sine overlaps of bin $b$ with the reference are
\begin{equation}
C_b\coloneqq\langle u,v_b\rangle,
\qquad
S_b\coloneqq\langle Ju,v_b\rangle.
\label{eq:coherence-overlaps}
\end{equation}
If all $c_1$ blocks undergo one common phase rotation, these overlaps vary as
$\cos\Delta_b$ and $\sin\Delta_b$. Their combined magnitude therefore defines
the coherence
\begin{equation}
\kappa_b
\coloneqq
\frac{\sqrt{C_b^2+S_b^2}}{\lVert v_b\rVert_2}
\in[0,1].
\label{eq:orbit-subspace-coherence}
\end{equation}
A value of one means that the phase-bin mean lies entirely within the
rotational plane spanned by $u$ and $Ju$; lower values indicate components
outside this common plane. Mean coherence across the seven non-reference bins
is $0.9927$ for \Winder{} and $0.3159$ for the control. This metric measures
whether the $c_1$ blocks rotate coherently, but not whether the phase bins occur
in the correct order around the loop. 

\begin{figure*}[!t]
  \centering
  \includegraphics[width=\textwidth]{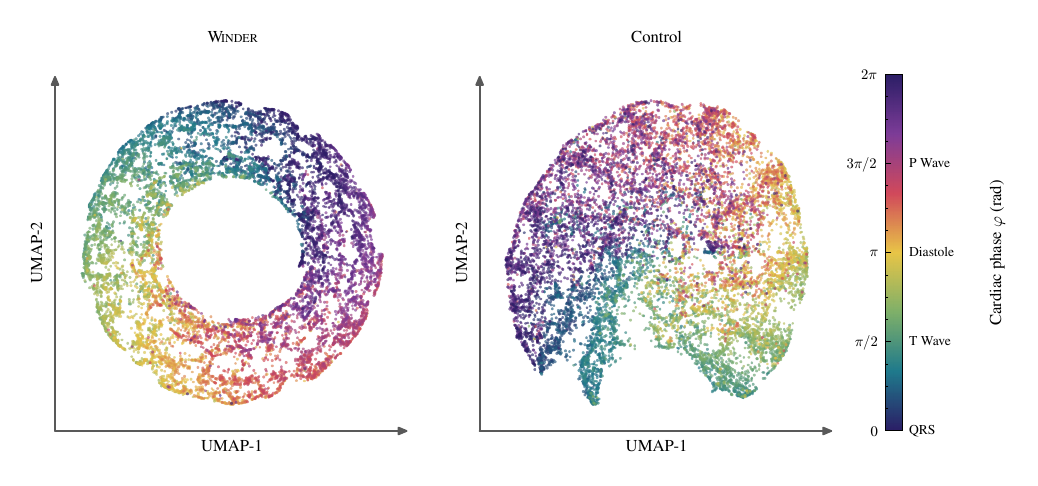}
  \caption{\textbf{UMAP visualisation of latent cardiac-phase organisation.}
  Token-level representations at seed~0 are shown separately for \Winder{} and
  the control. UMAP receives no cardiac-phase or diagnostic information, points
  are coloured post hoc by cardiac phase $\phi$.}
  \label{fig:umap-phase}
\end{figure*}

The trajectory in \cref{fig:phase-ring} complements this metric by showing that the bins also
follow their cardiac-phase order around the loop.
Tokens are grouped into 24 cardiac-phase bins, and each bin's mean
representation is projected onto the two-dimensional subspace in which the
rotation acts at the fundamental harmonic $n_{\mathrm{f}}{=}1$; the four
columns are the two \Winder{} seeds and the two control seeds of
\cref{app:training-runs}. In both \Winder{} runs the phase-binned trajectory
walks once around a closed circle in bin order; in both control runs it
collapses to a small path that neither closes nor preserves the ordering. The
figure and the table quantify
related but distinct properties on different splits: the RMS loop radius drawn
from folds~1--9 separates the arms by roughly a factor of seven, whereas the
sealed-fold amplitude separates them by a factor of five. The figure therefore
locates the mechanism, while the sealed-fold quantities measure it.

Phase organisation is also visible without the harmonic construction.
\cref{fig:umap-phase} embeds token-level representations with a general-purpose
non-linear projection that receives no phase, arm or diagnosis information, and
colours the result by $\phi$ afterwards. Each point is one token, an 80\,ms
slice of one recording, so colour organisation is a property of the
representation rather than of the projection: \Winder{}'s tokens form a
phase-ordered annulus and the control's do not. Both figures are encoded from
folds~1--9 at the reported evaluation cell and at the training seeds of
\cref{app:training-runs}, because the sealed fold did not persist per-record
latents; the sealed-fold amplitude and coherence values are those in
\cref{tab:bottomline}. The embedding is a qualitative illustration the geometry claim.

\subsection{Anomaly detection.}
One causal detector was fixed before unblinding and applied unchanged across
six synthetic perturbation families. Each maximum-severity result is read
against that cell's own zero-severity reference --- the same perturbation
applied at zero amplitude --- because these references differ between arms.
For ectopic beats at severity~2.0, \Winder{} reaches AUROC
$0.8662$, compared with $0.6198$ for the control. The reference-corrected
excesses are $+0.2758$ and $+0.1393$, and localisation hit rates are $0.779$ and
$0.025$. This is the battery's strongest detection result.

Latency is interpretable only when the detector first clears its own
zero-severity reference.
\Winder{} raises an alarm on $92.0\,\%$ of ectopic-beat records, whereas the
control alarms on $14.6\,\%$, so the control's 720\,ms median is conditioned
on a small, unrepresentative subset. The corresponding \Winder{} median is
80\,ms.

That median is exactly one token. The encoder partitions each recording into
125 non-overlapping 80\,ms patches (\cref{sec:method}), and the detector emits
one score per token, so an alarm can only be timestamped to a patch boundary
and no reported latency can fall below a single patch. The patching we elect
therefore sets the floor on this measurement: 80\,ms is the finest delay the
token grid can express, not a measured bound on how quickly the representation
responds to onset. Resolving faster detection would require a shorter patch,
and hence more tokens per record. A
different admissible detector for seed~0 on the same causal clock reaches
$0.916$ on this family. Selecting it per cell would have added roughly $+0.05$
and was excluded by the registered protocol.

The step~5{,}000 seed-averaged results for the remaining families bound the
detection claim. T-wave inversion gives a weak reference-corrected excess of
$+0.1031$ for \Winder{} against $+0.0250$ for the control. Amplitude attenuation
and lead dropout are effectively undetected, at $+0.0138$ and $+0.0115$ for
\Winder{}. Baseline wander reverses after reference correction, with the control
responding more strongly ($+0.2242$ against $+0.1240$). Together with the
ST-shift result, these measurements support no arm-discriminating claim for the
other four families.

%% file: 05_discussion.tex
\section{Discussion}
\label{sec:discussion}

\Winder{} uses a soft physiological prior: the transport loss encourages, but
does not force, representations to vary predictably with cardiac phase. The
expected phase action $R_\Delta$ is fixed analytically as a set of harmonic
rotations. Integer harmonics guarantee
that a complete cardiac cycle returns the latent coordinates to their starting
point, $R_{2\pi}=I$. Transport gain and latent-geometry measurements then
determine whether the learned representation actually follows this prescribed
action.

The matched comparison isolates the effect of phase transport; the control
differs only by disabling the transport loss.
Phase transport improves diagnostic accessibility while strengthening robustness to lead loss. The result is therefore not merely a better readout, but a performance gain accompanied by evidence that the intended cardiac-phase action organises the representation.

This declared geometry also provides a monitorable contract for a compact causal
encoder. 
Phase organisation and transport violations can be quantified directly from
the latent representations using metrics defined by the fixed harmonic action.
The anomaly results show the promise of this view for continuous monitoring at
the point of acquisition, including ambulatory telemetry and bedside rhythm
surveillance. 
These results suggest that phase-aware anomaly scores may support continuous monitoring, but they were evaluated only on synthetic perturbations and do not establish performance in a deployed clinical system.

This work adopts a single corpus and a deliberately simple encoder backbone to
give prominence to the method and stress-test the phase-transport mechanism in
isolation. The evidence is consequently a controlled within-study comparison
under a frozen linear readout, not a common-protocol leaderboard. 
Having established benefits against the control, future work should evaluate
broader corpora and stronger backbones, examine which physiological,
morphological, and diagnostic features are encoded by the invariant and
harmonic latent subspaces, and extend single-step latent prediction to
multi-step forecasts.

More broadly, cardiac phase is one instance of a general construction in which
a measured cyclic coordinate defines a fixed harmonic action on latent space,
allowing equivariant structure to be introduced without learning the symmetry
itself.

%% file: 06_conclusions.tex
\section{Conclusions}
\label{sec:conclusions}

\Winder{} introduces a phase-equivariant objective for physiological
representation learning. It represents cardiac phase through a fixed,
closed-form action on the latent space, whose invariant and phase-rotating
harmonic subspaces are specified without learnable transport parameters. The
transport objective encourages the learned representations to follow this
prescribed action, and the resulting organisation can be measured through
transport gain and latent-geometry diagnostics. Notably, because the construction
requires only a measured cyclic coordinate, \Winder{} opens a broader class of
phase-equivariant world models for cyclic biosignals, in which physiological
structure is injected analytically rather than purchased through parameter
scale.

%% file: 08_appendix_data.tex
%

\section{Data protocol}
\label{app:data}

\begin{table*}[!t]
  \centering
  \begin{tabular}{@{}c l p{0.62\textwidth}@{}}
    \toprule
    \# & Step & Operation and parameters \\
    \midrule
    1 & Read & The 16-bit waveform release at 500\,Hz is converted to physical
    millivolts in single-precision floating point as
    $(\mathrm{raw}-\mathrm{baseline})/\mathrm{gain}$, scaled by the declared
    unit factor of the recording header. Each signal's stored checksum is
    verified. \\
    2 & Verify extent & Each recording is required to contain 5{,}000 samples
    across 12 leads, every sample finite, before any filtering. A header that
    does not declare $f_s = 500$\,Hz is rejected. \\
    3 & Verify lead order & I, II, III, aVR, aVL, aVF, V1--V6, checked against
    each recording's own header. A mismatch is rejected rather than corrected
    by reordering. \\
    4 & Decimate & $500 \rightarrow 100$\,Hz by polyphase resampling with
    interpolation factor 1 and decimation factor 5: a single linear-phase
    Kaiser-windowed FIR filter, with group delay compensated internally, so
    output sample $n$ corresponds to input time $n/f_{\mathrm{out}}$ with no
    timing shift. The signal is extended at each endpoint by its boundary
    value rather than by zeros, so a non-zero baseline introduces no step
    discontinuity at $t = 0$ or $t = 10$\,s. \\
    5 & Window & The whole recording: 1{,}000 samples across 12 leads. No
    cropping, no stride, no overlap --- exactly one training example per
    recording. \\
    6 & Normalise & Per-lead $z$-score against corpus statistics fitted on
    the decimated signal over the training folds only
    (\cref{tab:norm-stats}). \\
    7 & Arrange layout & Lead-major, 12 leads by 1{,}000 samples, in
    single-precision floating point. \\
    \bottomrule
  \end{tabular}
  \footnotesize
  \caption{\textbf{ECG preprocessing pipeline.} }
  \label{tab:preproc-chain}
\end{table*}

\begin{table}[!t]
  \centering
  \footnotesize
  \setlength{\tabcolsep}{4pt}
  \begin{tabular}{
    l
    S[table-format=-1.3]
    S[table-format=1.3]
    l
    S[table-format=-1.3]
    S[table-format=1.3]
  }
    \toprule
    Lead & {$\mu$ [mV]} & {$\sigma$ [mV]} &
    Lead & {$\mu$ [mV]} & {$\sigma$ [mV]} \\
    \midrule
    I   & -0.002 & 0.171 & V1 &  0.000 & 0.234 \\
    II  & -0.001 & 0.167 & V2 & -0.001 & 0.338 \\
    III &  0.000 & 0.172 & V3 & -0.001 & 0.335 \\
    aVR &  0.002 & 0.143 & V4 & -0.002 & 0.311 \\
    aVL & -0.001 & 0.147 & V5 & -0.001 & 0.291 \\
    aVF & -0.001 & 0.147 & V6 & -0.001 & 0.243 \\
    \bottomrule
  \end{tabular}
  \caption{\textbf{Per-lead normalisation statistics.} Fitted corpus
  normalisation statistics. $\mu$ and $\sigma$ are the per-lead mean and standard deviation.}
  \label{tab:norm-stats}
\end{table}

This appendix states the preprocessing chain and phase-clock construction in
full, the normalisation convention and the evidence for choosing it, and the
composition of the pre-training pool against the sealed fold.

\Cref{tab:preproc-chain} gives the preprocessing steps.

All reported runs use a fixed per-lead normalisation: one mean and standard
deviation are fitted on the decimated training-fold signals and applied
unchanged to every recording. Unlike per-beat or per-recording normalisation,
this preserves between-recording amplitude variation within each lead. The values o
of the means and standard deviations are given in \cref{tab:norm-stats}.

\begin{table}[!t]
  \centering
  \footnotesize
  \setlength{\tabcolsep}{3pt}
  \label{tab:cohort-composition}
  \begin{tabular}{@{}l cc@{}}
    \toprule
    & Folds 1--9 & Fold 10 \\
    \midrule
    Recordings                        & 19{,}601 & 2{,}198 \\
    Patients                          & 16{,}965 & 1{,}904 \\
    \midrule
    Age, median (years)               & 61    & 63    \\
    Age, interquartile range          & 50--72 & 50--74 \\
    Age sentinel ($\mathrm{age}=300$) & 259   & 34    \\
    Male, \%                          & 52.2  & 51.5  \\
    \midrule
    Normal ECG (NORM), \%             & 43.6  & 43.8  \\
    Myocardial infarction (MI), \%    & 25.1  & 25.0  \\
    ST/T change (STTC), \%            & 24.0  & 23.7  \\
    Conduction disturbance (CD), \%   & 22.5  & 22.6  \\
    Hypertrophy (HYP), \%             & 12.2  & 11.9  \\
    \textsc{Unlabeled} recordings     & 371   & 40    \\
    \midrule
    Beats per recording, median       & 12    & 12    \\
    Phase yield, median               & 0.908 & 0.909 \\
    Recordings failing phase QC       & 202   & 20    \\
    \bottomrule
  \end{tabular}
  \caption{\textbf{Pre-training and sealed-fold cohort composition.}
  Diagnostic prevalences are multi-label and may sum to more than 100\%.
  \textsc{unlabeled} denotes recordings without an eligible diagnostic
  statement. The $\mathrm{age}=300$ sentinel denotes de-identified patients
  over 89. Phase yield is the fraction of samples assigned a phase. Phase-QC
  failures remain in the self-supervised pool but are excluded from
  phase-dependent analyses.}
\end{table}

\cref{tab:cohort-composition} the demographic and diagnostic composition of the pre-training folds and the test fold. All 19{,}601 fold-1--9 recordings are used for pre-training, and the normalisation statistics of \cref{tab:norm-stats} are fitted over the same 19{,}601. The 222 recordings that fail phase quality control (1.0\% of the corpus) are retained in the pool, since a model consuming no phase is
unaffected by them and filtering the pool on that criterion would drop
recordings for an unrelated reason.

%% file: 09_appendix_phase_assignment.tex

\section{Per-token phase assignment}
\label{app:phase-assignment}

Each 80\,ms token summarises an eight-sample patch of the decimated signal, so
a single scalar phase must stand for the whole patch. Two conventions are used,
for different purposes.

\emph{Centre convention (training).} Each token carries the phase at its patch
centre, 3.5 sample intervals after the patch start. Phase tags are computed
once per record offline and loaded alongside each waveform batch, so the
training loop never re-derives them.

\emph{Causal convention (reporting only).} Descriptors that must not read
beyond a token's own support --- time since the preceding R-peak, distance to
the nearest R-peak --- instead take the phase at the token's last sample. This
convention never enters training.

Assigning the causal boundary during training would impose a fixed 35\,ms
offset, which becomes a heart-rate-dependent fraction of the cycle: $0.056$\,rad
standard deviation across records at the fundamental, and $0.39$\,rad at the
highest retained harmonic, $n=7$. The error grows with harmonic order because
harmonic $n$ advances $n$ times faster in phase, so the tolerance on the token
timestamp is set by the highest retained harmonic rather than by the
fundamental.


%% file: training_appendix.tex
%

\section{Training specification}
\label{app:training}

\subsection{SIGReg quadrature}
\label{app:training-sigreg}

For all reported runs, $\mathcal{L}_{\mathrm{SIG}}$ is evaluated on the raw
projector tokens $z$, without rotation into a canonical phase frame
(\cref{app:phase-assignment}). Each evaluation uses $M=256$ random projection
directions. At each training step, the projection directions are sampled afresh from the unit sphere. 

For each axis, the characteristic-function integral in
\cref{eq:epps-pulley} is approximated using $J=17$ trapezoidal quadrature
knots over $[0,3]$.
The interval exploits the evenness of the integrand and therefore
represents the symmetric domain $[-3,3]$. The Gaussian weighting assigns
little mass beyond $|t|=3$, while retaining $J=17$ knots over the shorter
interval gives finer resolution near the origin, where the integrand varies
most. This choice is also supported by the LeJEPA ablation in Table~1a, where
$[-3,3]$ and $[-5,5]$ yield comparable linear-probe performance
\citep{balestriero2025lejepa}. The absolute scale of SIGReg depends on the quadrature grid.

Two alternative SIGReg frames were tested but not adopted. Demodulating each
token to phase zero (the ``canonical'' frame) produced transport gains from
$-0.38$ to $-0.17$ and reduced the effective rank to $2.9$--$11.2$. Reducing
each record to a canonical template (the ``record-canonical'' frame) instead
yielded zero transport gain because the representations concentrated in the
non-rotating invariant block. All reported runs therefore use the raw token
frame, with record-level SIGReg disabled.

\subsection{Augmentation provenance}
\label{app:training-augment}
\begin{table*}[t]
    \centering
    \begin{tabular}{@{}l p{0.31\textwidth} p{0.43\textwidth}@{}}
      \toprule
      Augmentation & Operation & Parameters \\
      \midrule
      Gaussian noise
        & Add independent Gaussian noise to the waveform.
        & Zero mean; standard deviation equal to $5\%$ of the record RMS. \\
      Narrow-band tone
        & Add a sinusoidal interference component.
        & Frequency: $25$\,Hz. Amplitude: uniform on $5\%$--$20\%$ of record
          RMS. Phase offset $\xi$: uniform on $[0, 2\pi)$. One shared tone,
          identical across leads. \\
      Low-frequency drift
        & Add a slowly varying baseline component.
        & Sum of $1$--$3$ sinusoids, count drawn uniformly from $\{1,2,3\}$;
          each component's frequency uniform on $0.05$--$0.4$\,Hz and amplitude
          uniform on $5\%$--$20\%$ of record RMS, with an independent phase
          offset $\xi$ uniform on $[0, 2\pi)$. Shared across all leads. \\
      Amplitude modulation
        & Multiply the waveform by a slowly varying amplitude envelope.
        & Envelope $1 + a\sin(2\pi f t + \xi)$. Depth $a$: uniform on
          $2\%$--$10\%$. Frequency $f$: uniform on $0.05$--$0.3$\,Hz. Phase
          offset $\xi$: uniform on $[0, 2\pi)$. One record-global envelope,
          identical across leads. \\
      Lead dropout
        & Set complete ECG leads to zero.
        & Number of dropped leads $k$: uniform on $\{1,2,3\}$. The $k$ leads
          are a uniformly random subset of the 12, drawn without replacement,
          independently per record. \\
      Per-lead gain
        & Scale each ECG lead independently.
        & Gain $g_l$: drawn independently and uniformly on $[0.7, 1.3]$ for
          each of the 12 leads, independently per record. \\
      \bottomrule
    \end{tabular}
    \footnotesize
    \caption{\textbf{Training-time augmentation settings}. Each transform is
    applied independently to each record with probability $0.5$, gated
    independently per augmentation type. All additive amplitudes are a
    fraction of that record's own clean-input RMS, computed once before any
    augmentation is applied, so a stacked dose always refers to the same
    physical scale.}
    \label{tab:training-augmentations}
  \end{table*}

Signal augmentation follows precedent from related time-series and ECG
models. Reverso includes augmentation in its zero-shot forecasting recipe,
although its ablation assigns a larger effect to synthetic pre-training data
\citep{fu2026reverso}. In the ECG setting, ER-JEPA independently applies
baseline wander, Gaussian noise, and power-line noise during pre-training
\citep{kim2026erjepa}. These works motivate signal augmentation generally, but
do not validate the specific six-transform stack used here.

The stack itself was selected in a preliminary architecture-scouting
experiment rather than from the four pre-training runs reported in
\cref{sec:training}. This experiment used one random seed, 5,000 training
steps, a 6,000-record training subset, and evaluation on fold~9. The full
augmentation stack achieved a macro-AUROC of $0.8589$, compared with $0.8513$
for the control arm. This provides preliminary support for
carrying the stack forward, but does not estimate uncertainty or isolate the
contribution of individual augmentations. See \cref{tab:training-augmentations} for the complete specification.

\subsection{Optimiser and schedule}
\label{app:training-optim}

\begin{table}[t]
    \centering
    \begin{tabular}{@{}lc@{}}
      \toprule
      Hyperparameter & Value \\
      \midrule
      Optimiser & AdamW \\
      Batch size & 64 \\
      Training steps & 30{,}000 \\
      Peak learning rate & $3\times10^{-4}$ \\
      Minimum learning rate & $1\times10^{-6}$ \\
      Warm-up steps & 5 \\
      AdamW $\beta$ & $(0.9,\,0.95)$ \\
      AdamW $\epsilon$ & $1\times10^{-8}$ \\
      Weight decay & $1\times10^{-4}$ \\
      Maximum gradient norm & 1.0 \\
      \bottomrule
    \end{tabular}
    \small
    \caption{\textbf{Optimiser and schedule hyperparameters}. Used for all reported
    pre-training runs.}
    \label{tab:training-optim}
  \end{table}

All optimiser and schedule settings were fixed across the four pre-training
runs and were not swept; their values are given in
\cref{tab:training-optim}. Training uses AdamW with a linear warm-up followed
by cosine learning-rate decay.

\subsection{Four training runs}
\label{app:training-runs}

The experiment follows a two-arm, two-seed design. The transport-enabled arm
uses $\lambda_{\mathrm{trans}}=1$, while the control retains the same cyclic
operator but assigns the transport loss zero weight. All remaining model,
data, and optimisation settings are identical across the four runs. Within
each seed, randomness is separated into independent streams for parameter
initialisation, mask sampling, SIGReg projections, augmentation, and data
order. The active stream states are stored in each checkpoint. Checkpoints are saved every 2,500 steps and at the end
of training.

\begin{table}[t]
  \centering
  \small
  
  \label{tab:training-runs}
  \begin{tabular}{@{}lccc@{}}
    \toprule
    Arm & Seed & $\lambda_{\mathrm{trans}}$ & Elapsed [min] \\
    \midrule
    Transport & 0 & 1.0 & 249.6 \\
    Transport & 1 & 1.0 & 254.9 \\
    Control   & 0 & 0.0 & 240.0 \\
    Control   & 1 & 0.0 & 225.3 \\
    \bottomrule
  \end{tabular}
  \caption{\textbf{Four pre-training runs.} Elapsed time is the wall-clock training
  time recorded for each run.}
\end{table}

Training was performed on a single NVIDIA A100 40\,GB GPU in two matched
concurrent pairs, with approximately 11.7\,GB peak memory per pair.

%% file: 07_appendix_ptbxl.tex
\section{Positioning within PTB-XL self-supervised learning}
\label{app:ptbxl}
\begin{table*}[t]
  \centering
  \footnotesize
  \setlength{\tabcolsep}{3.5pt}
  \definecolor{CompactParamShade}{HTML}{DDEED8}
  \definecolor{IntermediateParamShade}{HTML}{FBE3C2}
  \definecolor{LargeParamShade}{HTML}{F6DCDC}
  \tcbset{parameter badge/.style={
    on line,
    boxrule=0pt,
    arc=1.5pt,
    left=2pt,
    right=2pt,
    top=0pt,
    bottom=0pt
  },
  compact badge/.style={
    parameter badge,
    colback=CompactParamShade,
    colframe=CompactParamShade
  },
  intermediate badge/.style={
    parameter badge,
    colback=IntermediateParamShade,
    colframe=IntermediateParamShade
  },
  large badge/.style={
    parameter badge,
    colback=LargeParamShade,
    colframe=LargeParamShade
  }}
  \begin{tabular}{
    @{}
    p{0.14\textwidth}
    p{0.11\textwidth}
    p{0.19\textwidth}
    p{0.08\textwidth}
    p{0.23\textwidth}
    p{0.15\textwidth}
    @{}
  }
    \toprule
    Method &
    Parameters $[\mathrm{M}]$ &
    Pre-training examples &
    PTB-XL in pre-training? &
    PTB-XL evaluation protocol &
    AUROC \\
    \midrule
    \multicolumn{6}{l}{\textcolor{CasalBrand}{\textit{Latent-prediction objectives}}} \\
    \midrule

    \Winder{} &
    1.2 &
    \(17\mathrm{k}\) &
    \(\checkmark\) &
    100\,Hz; full 10\,s record; frozen representation with a plain linear
    probe; folds 1--8 train and fold 10 test &
    0.8702 \\
    \addlinespace[2pt]

    S4-JEPA \citep{almasud2026ecgscaling} &
    \tcbox[intermediate badge]{3.0} &
    \tcbox[large badge]{\(11\mathrm{M}\)\textsuperscript{a}} &
    \(\times\) &
    240\,Hz; frozen mean-pooled linear evaluation; concrete PTB-XL fold
    assignment not stated &
    0.880 \\
    \addlinespace[2pt]

    LeNEPA \citep{chemeris2026lenepa} &
    \tcbox[intermediate badge]{\emph{3.6}} &
    N.A. &
    \(\checkmark\) &
    5,000-sample records; fixed-horizon pre-training; frozen
    intermediate-layer probe; label mapping insufficiently specified &
    0.880\textsuperscript{c} \\
    \addlinespace[2pt]

    JEPA ViT-XS \citep{weimann2025jepaecg} &
    \tcbox[intermediate badge]{3.6} &
    \tcbox[large badge]{\(1.0\mathrm{M}\)} &
    \(\checkmark\) &
    500\,Hz; 2.5\,s crops with record-level probability averaging; frozen
    encoder with learned cross-attention pooling &
    0.915 \\

    JEPA ViT-XS &
    \tcbox[intermediate badge]{3.6} &
    \tcbox[large badge]{\(0.80\mathrm{M}\)} &
    \(\times\) &
    Same protocol as the all-corpora JEPA row &
    0.915 \\

    JEPA ViT-XS &
    \tcbox[intermediate badge]{3.6} &
    \tcbox[compact badge]{\(17\mathrm{k}\)} &
    \(\checkmark\) &
    Same protocol as the all-corpora JEPA row &
    0.912 \\
    \addlinespace[2pt]

    ECG-JEPA \citep{kim2024ecgjepa} &
    \tcbox[large badge]{87.2} &
    \tcbox[large badge]{\(0.17\mathrm{M}\)} &
    \(\times\) &
    Eight leads at 250\,Hz; full 10\,s record; frozen mean-pooled
    representation with linear classifier; official folds &
    0.912 \\
    \addlinespace[2pt]

    CardioState-JEPA \citep{shafiq2026cardiostatejepa} &
    \tcbox[large badge]{\emph{94.3}} &
    \tcbox[large badge]{\(>5.9\mathrm{M}\)\textsuperscript{b}} &
    \(\times\) &
    Cross-modal ECG--PPG--PCG pre-training; frozen linear probe;
    official-style PTB-XL split &
    0.891 \\

    \addlinespace
    \multicolumn{6}{l}{\textcolor{SiennaBrand}{\textit{Contrastive objectives}}} \\
    \midrule

    S4-CPC \citep{almasud2026ecgscaling} &
    \tcbox[intermediate badge]{3.0} &
    \tcbox[large badge]{\(11\mathrm{M}\)\textsuperscript{a}} &
    \(\times\) &
    Same protocol and split limitation as S4-JEPA &
    0.860 \\
    \addlinespace[2pt]

    Poly-Window \citep{yuan2025polywindow} &
    \tcbox[large badge]{\emph{3.8}} &
    \tcbox[compact badge]{\(22\mathrm{k}\)} &
    \(\checkmark\) &
    100\,Hz; full 10\,s record; frozen encoder with linear classifier;
    official downstream folds &
    0.891 \\
    \addlinespace[2pt]

    ECG-CPC \citep{mehari2021ssl} &
    \tcbox[large badge]{5.8} &
    \tcbox[large badge]{\(55\mathrm{k}\)} &
    \(\checkmark\) &
    100\,Hz; frozen linear evaluation with 2.5\,s crop aggregation;
    all 71 statements &
    N.A. \\

    \bottomrule
  \end{tabular}

  \vspace{3pt}
  \begin{minipage}{\textwidth}
    \scriptsize
    \textsuperscript{a}The S4 studies report pre-training samples rather than
    complete ECG records. \textsuperscript{b}CardioState-JEPA combines ECG
    records, PPG segments, PCG recordings, and paired multimodal corpora; the
    paired-corpus counts are not reported. \textsuperscript{c}LeNEPA's reported
    AUROC uses all 71 PTB-XL labels rather than the five diagnostic
    superclasses.
  \end{minipage}
  \caption{\textbf{Published self-supervised comparisons on PTB-XL.}
  Results are reproduced in each source's original evaluation setting and are
  not a common leaderboard. AUROC values refer to the five diagnostic
  superclasses. Green, orange, and red shading in the parameter and
  pre-training-example columns denotes
  \protect\tcbox[compact badge]{\(0.5\text{--}1.5\times\)},
  \protect\tcbox[intermediate badge]{\(>1.5\text{--}3\times\)}, and
  \protect\tcbox[large badge]{\(>3\times\)} \Winder{} values, respectively. Despite
  differences in evaluation protocols, \Winder{} achieves AUROC values in the
  same broad range as the selected models while using fewer parameters and less
  pre-training data.}
  \label{tab:ssl-comparisons}
\end{table*}

This appendix situates \Winder{} (\(1.2\mathrm{M}\) parameters) within the published
self-supervised ECG representation-learning literature on PTB-XL. The
comparison is restricted to methods that report PTB-XL diagnostic performance
and provide sufficient information about their pre-training data, model scale,
and downstream evaluation protocol. The resulting set is curated, not
exhaustive: its purpose is to identify the nearest methodological and empirical
reference points for \Winder{}, not to construct a leaderboard. Despite its
ubiquity in the literature, there is no shared protocol for PTB-XL evaluation.

LeNEPA (\emph{3.6}\(\mathrm{M}\)\footnote{Italicised parameter counts are
estimates rather than values explicitly reported by the source. This count is
estimated from the released PTB-XL ViT-XS configuration.}) provides the nearest
methodological comparison. Both
methods combine causal latent prediction with SIGReg, however there are two key
differences in how embedded windows are related: LeNEPA applies SIGReg to the
sequence of token embeddings from each ECG, preventing collapse without giving
their arrangement a cardiac-phase meaning; \Winder{} includes the equivariant
transport loss that requires latents to follow measured cardiac-phase.

Beyond LeNEPA, the principal signal-only JEPA references are S4-JEPA
(\(3.0\mathrm{M}\)) and the models of \citet{weimann2025jepaecg} and
\citet{kim2024ecgjepa}.
Weimann JEPA is evaluated at several scales; here, we consider only the smallest, ViT-XS (\(3.6\mathrm{M}\)), as the closest scale comparison to \Winder{}.
It uses masked prediction, is pre-trained on configurations ranging from PTB-XL alone to ten ECG corpora, and evaluates frozen representations through learned cross-attention pooling over 2.5\,s crops. 
ECG-JEPA (\(87.2\mathrm{M}\)) instead uses a substantially larger Transformer with cross-pattern attention,
pre-trained on 174,140 external ECGs and evaluated through mean-pooled frozen
representations. 
Neither method assigns cardiac phase an explicit action in
the latent space.

Readout capacity is itself a material protocol difference. Weimann's learned
cross-attention pooling can adaptively combine crop-level features, whereas
\Winder{} uses parameter-free mean pooling followed by a linear classifier.
\Cref{tab:ssl-comparisons} therefore does not support a head-to-head accuracy
claim. The narrower result is that phase-transport training yields diagnostic
information that remains accessible to a deliberately low-capacity readout,
without a learned sequence aggregator. This is directly relevant to deployment
because the downstream readout adds only a linear classifier.

Contrastive predictive objectives provide a complementary set of self-supervised references.
S4-CPC (\(3\mathrm{M}\)) of \citet{almasud2026ecgscaling} is compact but
is pre-trained on 11.2 million external ECG samples, and its reported PTB-XL
split is insufficiently specified for direct comparison. ECG-CPC
(\(5.8\mathrm{M}\)) is similarly compact, but reports the
all-71 task and does not clearly restrict PTB-XL pre-training to the downstream
training folds. Poly-Window \citep{yuan2025polywindow}
(\emph{3.8}\(\mathrm{M}\)\footnote{Estimated from the canonical 1D
ResNet18-512/128 architecture.}) most closely matches \Winder{}'s 100\,Hz,
full-record frozen-probe evaluation, although self-supervised pre-training
folds are not reported.

CardioState-JEPA \citep{shafiq2026cardiostatejepa}
(\emph{94.3}\(\mathrm{M}\)\footnote{Estimated from the released configuration
at repository commit \texttt{ab242c3}.}) provides a separate
cross-modal reference, learning a shared representation from ECG, PPG, and PCG
with cardiac phase included as an auxiliary prediction target. This differs
from \Winder{}, which uses ECG alone and imposes phase as a fixed group action
on the latent space rather than as a quantity to predict.
Their ablations independently suggest that cardiac-phase supervision is useful, although phase is treated as a prediction target rather than an equivariant constraint.
CardioState-JEPA is therefore relevant to cardiac-state representation, but
its multimodal pre-training prevents a direct signal-only comparison.

Although none of the published results is directly comparable with
\Winder{} under a fully matched protocol, its diagnostic performance lies
within the same broad range as the selected self-supervised models.
\Winder{} reaches this range with a \(1.2\mathrm{M}\)-parameter model,
pre-training restricted to 19,601 PTB-XL records, and parameter-free mean
pooling followed by a linear classifier. Several reference methods use more
expressive learned readouts, including Weimann's cross-attention pooling.
\Cref{tab:ssl-comparisons} therefore supports a narrower,
deployment-relevant claim: phase-transport training produces representations
from which diagnostic information is linearly accessible without a learned
sequence aggregator.
